\documentclass[lettersize,journal]{IEEEtran}
\usepackage[caption=false,font=normalsize,labelfont=sf,textfont=sf]{subfig}
\usepackage{textcomp}
\usepackage{stfloats}
\usepackage{url}
\usepackage{verbatim}
\usepackage{graphicx}
\usepackage{cite}  % or \usepackage{natbib}, but not both
\usepackage{times}
\usepackage{latexsym}
\usepackage{amsmath}
\usepackage{booktabs}
\usepackage{tabularray}
\usepackage{array}
\usepackage{ragged2e}
\usepackage{multirow}
\usepackage{amsfonts}
\usepackage{colortbl}
\usepackage[ruled,vlined]{algorithm2e}
\usepackage{setspace}
\usepackage[english]{babel}
\usepackage{wasysym}
\usepackage{longtable}

\begin{document}

\title{Retrieval-Augmented Mixture of LoRA Experts for Uploadable Machine Learning}

\author{Ziyu~Zhao, Leilei~Gan, Guoyin~Wang, Yuwei~Hu, Tao~Shen, Hongxia Yang,~\IEEEmembership{Member,~IEEE}, Fei~Wu,~\IEEEmembership{Senior~Member,~IEEE}, Kun~Kuang\IEEEauthorrefmark{2}% <-this % stops a space
\thanks{\IEEEauthorrefmark{2}{Corresponding author.}}

\IEEEcompsocitemizethanks{
\IEEEcompsocthanksitem Z. Zhao, L. Gan, Y. Hu, T. Shen F. Wu and K. Kuang are with the College of Computer Science and Technology, Zhejiang University, China.
\protect
(E-mail: benzhao.styx@gmail.com; leileigan@zju.edu.cn; 
hyw-luv@zju.edu.cn;
tao.shen@zju.edu.cn;
wufei@zju.edu.cn;
kunkuang@zju.edu.cn).

\IEEEcompsocthanksitem G. Wang and H. Yang are with ByteDance Inc., Seattle, USA.
\protect
(E-mail: guoyinwang.duke@gmail.com;
hx.yang@bytedance.com).
}
}

% \IEEEpubid{0000--0000/00\$00.00~\copyright~2021 IEEE}
% Remember, if you use this you must call \IEEEpubidadjcol in the second
% column for its text to clear the IEEEpubid mark.

\maketitle

\begin{abstract}
Low-Rank Adaptation (LoRA) offers an effective solution for fine-tuning large language models (LLMs). Its modular nature allows the integration of diverse domain-specific LoRAs to enhance LLM capabilities. The emergence of open-source platforms like Huggingface has given rise to a new computational paradigm termed Uploadable Machine Learning (UML). In this framework, edge-side contributors train specialized LoRA adapters and then upload them to a central platform that leverages their plug-and-play nature to provide personalized services for downstream heterogeneous requests. Previous research on LoRA composition has centered on isolated tasks or static adapter selection during training, which does not adequately address the dynamic and expanding pool of LoRAs in UML nor sufficiently meet the needs of downstream heterogeneous requests.  To bridge this gap, we propose Retrieval-Augmented Mixture of LoRA Experts (RAMoLE), a retrieve-then-compose framework that adaptively retrieves and composes multiple LoRAs according to the input prompts. RAMoLE contains three main components: firstly, identifying and retrieving LoRAs relevant to the given input through LoraRetriever; secondly, coordinating various dynamically retrieved LoRAs through a carefully designed on-the-fly MoLE mechanism; and thirdly, developing efficient batch inference to accommodate heterogeneous requests. The experimental results indicate that RAMoLE consistently outperforms the baselines, emphasizing its effectiveness and flexible scalability.
\end{abstract}


\section{Introduction}

\begin{figure}[t]
    \centering
    \includegraphics[width=\linewidth]{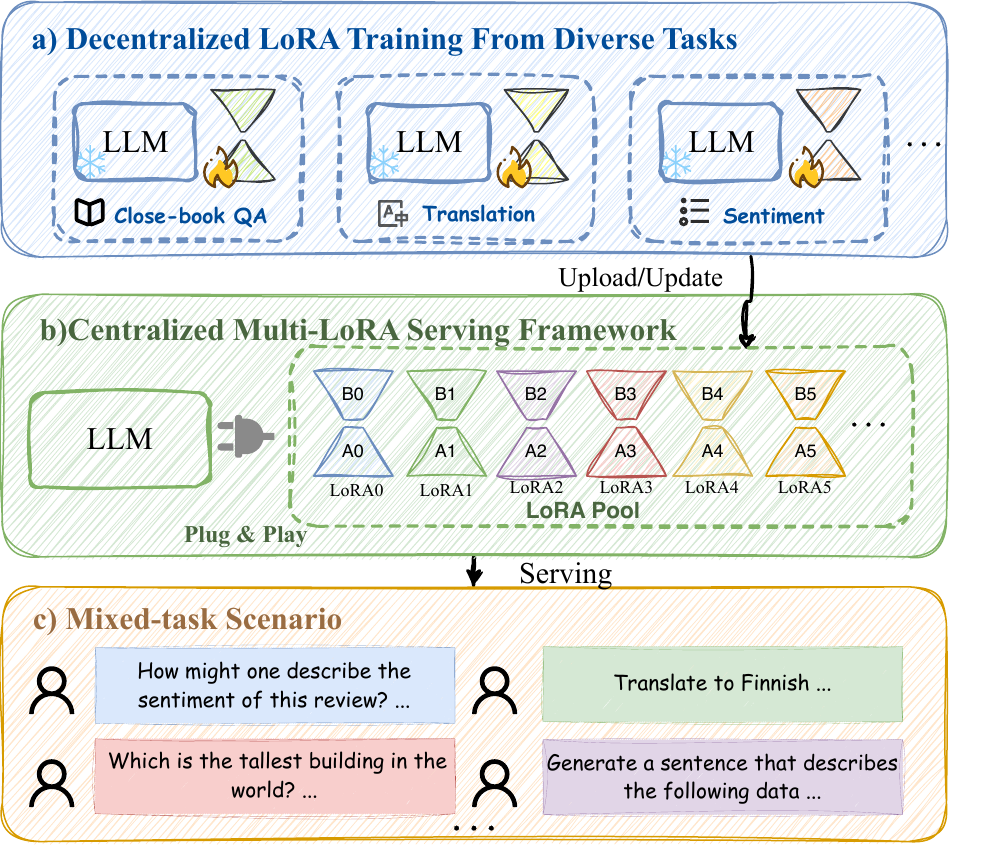}
    \caption{Illustration of UML. a) LoRAs from various domains and tasks aimed at enhancing specific capabilities of the LLM can be uploaded to or updated to the LoRA pool. b) The multi-LoRA serving framework aims to leverage the plug-and-play nature of LoRAs to offer comprehensive services. c) The downstream tasks, presented in a mixed-task form, require personalized expert routing.}
    \label{fig:mix_task}
\end{figure}

% 大模型在不同的领域获得了成功
\IEEEPARstart{R}ecently, leveraging the principles of scaling laws~\cite{kaplan2020scaling}, Large Language Models (LLMs) like openAI's GPT-3.5~\cite{ouyang2022training}, GPT-4~\cite{achiam2023gpt}, Meta's LLama~\cite{touvron2023llama} and many other Foundation Models~\cite{wei2021finetuned,chowdhery2023palm,bai2023qwen,young2024yi} have achieved notable success across various natural language processing (NLP) tasks~\cite{hadi2023survey,wang2023survey}. However, these models, primarily trained on diverse internet datasets, sometimes struggle in specialized areas. To address this, numerous studies~\cite{liu2023moelora,yang2023fingpt,hadi2023survey} have explored fine-tuning LLMs using domain-specific data, thereby adapting them to fulfill specific needs. Due to the prohibitively high computation costs for fine-tuning LLMs on specific domains, there is a growing shift towards Parameter-Efficient Fine-Tuning (PEFT)~\cite{liu2022p, hu2023llm, hu2021lora}, which only updates a small fraction of the model's parameters or integrates new trainable parameters that augment the model's capabilities. 

% As the focus on open-source communities like Hugging Face, ModelScope, and Civit AI grows, there is a notable trend of sharing an increasing number of PEFT modules and with LoRA emerging as the
% dominant finetuning approach, trained on a variety of domains and tasks, across these platforms.
As interest grows in open-source communities like Hugging Face\footnote{https://huggingface.co/}, ModelScope\footnote{https://modelscope.cn/home}, and Civit AI\footnote{https://civitai.com/}, there is an increasing trend towards sharing numerous PEFT modules trained on a broad spectrum of domains and tasks across these platforms. Within these PEFT methods, LoRA has become the dominant fine-tuning method~\cite{luo2024stylus}.  LoRA achieves efficiency by decomposing the weight matrices of a pre-trained model into low-rank factors, allowing it to adapt the model with minimal parameter changes. This approach not only significantly reduces computational load but also maintains or improves model performance through targeted updates. Its modularity and plug-and-play capabilities make it highly adaptable for various tasks and applications. This trend is giving rise to a new decentralized computing paradigm, which we refer to as \textit{Uploadable Machine Learning} (UML)~\cite{zhao2024loraretriever,luo2024stylus,liu2024lora}. 
As shown in Fig.\ref{fig:mix_task}, this new computing paradigm involves three components: 
a) The edge-side domain-specific LoRA contributors. These contributors train plug-and-play, domain-specific LoRA modules in a decentralized manner using local, domain-specific data. The trained LoRA parameters are then uploaded to a centralized service platform.
b) Centralized Multi-LoRA Serving platform. The server side runs an LLM equipped with a dynamic pool of LoRAs. The purpose of the LoRA pool is to enhance the LLM's capabilities across various domains, allowing it to deliver strong performance in a wider range of fields. The LLM leverages the plug-and-play capability of LoRA to dynamically load various LoRAs, allowing it to deliver personalized services tailored to diverse downstream requests.
c) Mixed-task Downstream Requests. The downstream requests cover a diverse range of tasks and are in mixed-task form. The heterogeneous nature of these requests requires that the server side utilize the appropriate LoRA for different prompts to deliver personalized services.

% Challenges in Uploadable Machine Learninig
The serving system for UML, designed to equip LLMs with a large pool of LoRAs for UML scenarios, faces two primary challenges. First, as the pool of LoRAs expands and evolves, developing efficient routing mechanisms for this dynamically changing collection becomes critical. This includes the need for zero-shot routing strategies specifically for newly uploaded LoRAs, ensuring they are integrated smoothly without prior adaptation. Second, the nature of downstream tasks in this paradigm typically involves a mixed-task format, which necessitates personalized routing of specific LoRAs for each input prompt. Unlike a one-size-fits-all approach, this requires a tailored solution that dynamically coordinates LoRA modules from various domains to effectively respond to the diverse and specific demands of user prompts. Platforms like ChatGPT and Gemini exemplify this scenario, highlighting the complexity of providing accurate and contextually appropriate responses in real time.

Recently, some work has focused on how to compose different LoRAs to enhance the model's capabilities. Several research has explored the integration of the mixture of expert (MoE;~\cite{jacobs1991adaptive,jordan1994hierarchical}) with LoRAs~\cite{wang2022adamix,liu2023moelora,zadouri2023pushing,muqeeth2023soft,wu2023mole}.
However, these methods fix the selection of LoRAs during training, lacking the flexibility to dynamically update and scale in scenarios where the LoRA pool may consistently expand in the UML scenario. This rigidity violates the inherently dynamic and uploadable nature of decentralized training involving domain-specific LoRA experts. LoRAHub~\cite{huang2023LoRAHub} and AdapterSoup~\cite{chronopoulou2023adaptersoup} explore composing LoRAs for specific downstream tasks.
However, these two methods provide a one-size-fits-all approach to downstream tasks, failing to accommodate the heterogeneous nature of diverse requests in a UML setting.

In this paper, we introduce a Retrieval-Augmented Mixture of Experts (RAMoLE), a retrieve-then-compose framework designed to exploit the plug-and-play nature of LoRA for UML. Our framework consists of three key components: (1) \textbf{Input-aware LoRA Retrieval:} The first step of our framework is aligning user inputs with the corresponding LoRAs through sentence embeddings and is further refined by an instruction fine-tuning~~\cite{su2022one,asai2022task} for effective LoRA retrieval. 
%Through the retriever, we achieve personalized routing in an off-the-shelf manner without training parameters of LLM, yielding good results in retrieving out-of-domain LoRAs.
Through the retriever, we achieve a more flexible LoRA routing mechanism, whose training stage is disentangled from the training and inference of the LLM.
(2) \textbf{On-the-fly Mixture of LoRA Experts:}
After retrieving the corresponding LoRAs for each sample, we need to finely differentiate and compose these LoRAs for targeted generation. To this end, we propose a novel MoLE mechanism that involves training an additional RouterLoRA, which can be seamlessly integrated into the LLM in an off-the-shelf manner. This mechanism uses an attention mechanism to determine the weights corresponding to the retrieved LoRAs, thereby achieving a more flexible and dynamic MoLE architecture.
It is worth noting that vanilla LoRA composition methods, which involve either averaging parameters or outputs, assign uniform weight to all loaded LoRAs. Such approaches may not allocate the most effective weights to the most suitable LoRA, potentially leading to suboptimal outcomes. In contrast, our proposed MoLE mechanism, utilizing a RouterLoRA, finely tunes the allocation of weights to each LoRA, dynamically differentiating each one to enhance overall performance
(3) \textbf{Batch Inference of Multiple LoRAs:} Most previous work on the input-adaptive inference of LLMs does not support batch inference~\cite{zhou2020bert, chronopoulou2023adaptersoup}. To tackle the challenge of heterogeneous batched requests, we construct a unique LoRA mapping matrix for batch samples. This allows for tailored inferences through efficient matrix multiplication, ensuring each request activates its corresponding LoRAs while maintaining batch processing efficiency.

To assess the performance of RAMoLE, we established a UML evaluation benchmark comprising 48 LoRAs spanning a variety of natural language understanding and generation tasks. The experimental results underline the effectiveness of the proposed methods in serving both in-domain and out-of-domain downstream requests. Furthermore, the retrieval routing method demonstrates robust generalization capabilities: despite the retriever being trained on only 40\% of the tasks, it effectively identifies the appropriate LoRAs for unseen tasks. Similarly, the RouterLoRA, also trained on just 40\% of the tasks, successfully routes LoRAs for tasks it has not previously encountered, achieving commendable performance.

We note that a shorter conference version of this paper appeared in~\cite{zhao2024loraretriever}. Our conference version of the paper did not explicitly define the entire scenario. This manuscript more thoroughly and comprehensively defines the setting as \textit{Uploadable Machine Learning}. Additionally, the conference version simply applied the same weights for composing the retrieved LoRAs, an approach that may not allocate the most effective weights to the most appropriate LoRA, potentially resulting in suboptimal outcomes. In this manuscript, we introduce a novel on-the-fly MoLE method that employs a RouterLoRA to distinguish between each retrieved LoRA, enhancing performance in the UML scenario. Additionally, to accommodate the new MoLE mechanism, the batch inference mechanism has been adapted to facilitate a more efficient inference process.

Our contribution can be summarized as follows.
\begin{itemize}
    \item We explore a new computing paradigm termed Uploadable Machine Learning, which entails managing a continuously expanding LoRA pool for serving heterogeneous downstream requests.
    \item We propose RAMoLE, a novel framework designed for the massive deployment of multiple LoRAs in response to personalized downstream requests. This framework consists of input-aware LoRA retrieval processes, strategic LoRA combinations, and a customized batch inference mechanism.
    \item We propose a novel MoE method that leverages a RouterLoRA as the gating function and assigns weights to each proposed LoRA through an attention mechanism. This approach enables RAMoLE to achieve more flexible and on-the-fly MoE, dynamically accommodating the retrieved LoRAs and generalizing to unseen LoRAs as a zero-shot router.
    \item Empirically, we show that our proposed RAMoLE framework outperforms other baselines in mixed-task scenarios. Moreover, LoraRetriever and the on-the-fly MoE mechanism achieve good generalization on unseen tasks and LoRAs as zero-shot routers. 
\end{itemize}

\section{Related Work}
\subsection{Mixture of Experts} 
The Mixture of Experts (MoE) method combines various specialized sub-modules, guided by a gating network to tailor responses to different input types~\cite{jacobs1991adaptive,jordan1994hierarchical,shen2023mixture,riquelme2021scaling,dou2023loramoe}. 
Some work~\cite{wang2022adamix, zadouri2023pushing,zhu2023sira, liu2023moelora,dou2023loramoe,ostapenko2023case,feng2024mixture} focuses on using the MoE method for PEFT to achieve more effective and efficient model fine-tuning.
Other work~\cite{wu2023mole,muqeeth2023soft} focuses on the use of MoE to post hoc coordinate existing LoRA experts without specifically training the experts' parameters.
Specifically, AdaMix~\cite{wang2022adamix} introduces a mixture-of-adaptations approach for fine-tuning models efficiently by combining multiple adaptation methods to reduce parameter overhead.
\cite{ostapenko2023case} demonstrates the efficacy of combining MoE with lightweight parameter-efficient tuning methods to enhance instruction tuning for large language models, achieving performance comparable to full fine-tuning while significantly reducing computational resource requirements.
\cite{zadouri2023pushing} enhances MoE by developing a highly parameter-efficient model that achieves performance on par with full fine-tuning while significantly reducing parameter usage.
SiRA~\cite{zhu2023sira} introduces a sparse mixture approach to low-rank adaptation, improving the efficiency and performance of large language models by optimizing sparse computation and reducing overfitting.
MoELoRA~\cite{liu2023moelora} leverages MoE and LoRA to efficiently fine-tune large language models for diverse medical tasks, addressing the issues of task variety and high computational cost.
LoRAMoE~\cite{dou2023loramoe} integrates MoE and LoRA to enhance large language models' capability in maintaining and applying world knowledge efficiently, significantly improving model performance while reducing computational requirements.
Mixture-of-LoRAs~\cite{feng2024mixture} presents a method that leverages a mixture of LoRA modules to enhance the multitask performance of large language models, achieving significant efficiency in fine-tuning across various tasks.
These works focus on integrating MoE and LoRA for more efficient fine-tuning of LLMs and reducing computational resources during inference. Our work is orthogonal to these approaches, as we investigate post-hoc LoRA combination, aligning with recent trends in adapting a large LoRA pool for various downstream tasks.

MoLE~\cite{wu2023mole} proposes an inference-time MoE method to enhance the parameter efficiency and flexibility of large language models, allowing for improved fine-tuning and adaptation across diverse tasks without significant computational overhead.
SMEAR~\cite{muqeeth2023soft} improves the performance of mixture-of-experts (MoE) models by using a soft merging approach for expert parameters combined with adaptive routing, avoiding the inefficiencies of discrete routing decisions and resulting in better expert specialization and overall model efficiency.
PATGOOSE~\cite{muqeeth2024learning} explores a framework for zero-shot learning where a model learns to dynamically route inputs among specialized experts, thereby achieving generalization to unseen tasks by leveraging the expertise of various pre-trained modules, significantly enhancing the model's adaptability and performance in zero-shot scenarios.
\cite{ostapenko2024towards} presents a modular approach for LLMs by constructing and reusing a library of LoRA modules, enabling efficient fine-tuning and adaptation for multiple tasks, thereby facilitating the creation of versatile and resource-efficient LLMs that can be easily customized for different applications
These methods necessitate that the router has encountered all LoRAs during training and requires the selection of LoRAs to be fixed at that time. This limitation prevents these methods from managing the continuously expanding LoRA pool in UML and from routing for unseen LoRAs.

\subsection{Adapter Merging}
In addition to model ensembling through the MoE, there is an increasing focus on aggregating adapters from different domains through the method of Adapter Merging. 
\cite{zhang2023composing} proposes to compose these parameter-efficient modules through linear arithmetic operations in the weight space to integrate different module capabilities.
AdapterSoup~\cite{chronopoulou2023adaptersoup} aggregates different adapters in the parameter space, allowing large language models to adapt to new domains without additional training. 
LoRAHub~\cite{huang2023LoRAHub} employs random sampling of LoRA parameters from various domains and tasks, followed by black-box optimization to learn the weights of different LoRA parameters without involving model gradient backpropagation. 
LoRA-Flow~\cite{wang2024lora} dynamically fuses multiple Low-Rank Adaptation (LoRA) modules to enhance the generative capabilities of LLMs, enabling more efficient and effective task-specific fine-tuning without substantial computational overhead.
$\pi$-Tuning ~\cite{wu2023pi} considers not only language model experts but also visual model experts. 
It obtains weights for different LoRAs through task embeddings for downstream tasks, similarly utilizing weighted averaging in the parameter space for targeted model parameter adaptation. 
LoRA-Composer~\cite{yang2024lora} proposes a framework that utilizes Low-Rank Adaptation to enable multi-concept customization in training-free diffusion models, addressing the challenge of integrating multiple concept LoRAs for complex and varied image generation tasks.
\cite{zhong2024multi} explores Multi-LoRA Composition for combining multiple LoRA modules to improve the quality and efficiency of image generation, allowing for the seamless integration of various image attributes and styles.

These methods offer a one-size-fits-all solution for downstream tasks, which is not suitable for mixed-task scenarios that require personalized services. Moreover, adapter merging methods may encounter issues with parameter interference~\cite{tang2024merging}, which can severely degrade the performance of the merged model.

\subsection{Personalized LoRA serving}
\cite{sheng2023s} propose S-LoRA to discuss serving thousands of concurrent LoRA. The framework targets scenarios in which multiple tasks must be handled simultaneously without compromising the efficiency of the base models. \cite{wen2023batched} propose FLoRA, which enables efficient batching of diverse request types in the LoRA of foundation models. These studies discuss how to deploy or train personalized LoRAs. However, these methods can only utilize a single user-specified LoRA during inference, failing to fully leverage the combination of LoRAs from different tasks. Moreover, the primary focus of these discussions is on computational strategies in GPUs and training strategies, which are orthogonal to the routing strategies with which we are concerned.

% \subsection{Federated Learning}
% Federated Learning (FL)~\cite{mcmahan2017communication} is a distributed machine learning framework that enables multiple participants to collaboratively train a model without directly exchanging their data, thereby enhancing data privacy and security. Several recent studies have discussed the application of federated learning to pre-trained language models, particularly leveraging PEFT techniques to reduce communication costs~\cite{bai2024federated, chen2023federated, ye2024openfedllm, zhang2024towards}. The difference between our UML scenario and FL is that the number of edge devices in our scenario is not fixed, and neither continuous communication nor ongoing model training is required. Instead, it more closely mirrors real-world scenarios where different contributors upload models from various domains to the cloud platform and perform post-hoc model fusion to provide comprehensive services.

\section{Preliminaries}
This section begins with a concise introduction to the Low-Rank Adaptation and Mixture of LoRA Experts, followed by a detailed formalization of our \textit{Uploadable Machine Learning} setting.

\subsection{Low-Rank Adaptation}
Directly fine-tuning large language models with all parameters is computationally intensive and is not feasible in low-resource scenarios. Based on the idea that only a small number of low-rank parameters need to be fine-tuned for sufficient performance in new domains, \cite{hu2021lora} proposed the Low-Rank Adaptation, where the LoRA module can be combined with the pre-trained parameters in parallel for efficient inference. 

Specifically, given pre-trained weights $W_0 \in \mathbb{R}^{d\times k}$ of a sub-module of LLM, the LoRA adds an extra trainable weight matrix as $W_0 + \Delta W = W_0 + BA$, where $\Delta W$ can be decomposed into two smaller matrices $B\in \mathbb{R}^{d\times r}$ and $A \in \mathbb{R}^{r \times k}$, where $r$ stands for the rank of $\Delta W$ and the rank $r \ll min(d,k)$. The forward pass for a layer $x^{\prime} = W_0 x$ can be modified as follows:
\begin{equation}
    x^{\prime} = W_0 x+\Delta Wx = W_0 x + BAx,
\end{equation}
where $x \in \mathbb{R}^d$ is the input and the $x^{\prime} \in \mathbb{R}^d$ denote the output.

\subsection{Mixture of LoRA Experts}
The mixture of LoRA Experts (MoLE) is an efficient ensemble approach that enhances model scalability by selectively activating only a subset of parameters, thus increasing the model's capacity without incurring additional computational costs. Recently, several studies have explored the integration of Low-Rank Adaptation (LoRA) with MoE for diverse applications. Unlike traditional transformer models, each MoE layer with the original parameter $W$ consists of $k$ independent LoRA experts $\{\Delta W_i\}_{i=1}^k$, along with a gating function $G(\cdot)$ that models the probability distribution to determine the weighting of the outputs of these experts. The output of a MoLE layer can be given by:
\begin{equation}
    x^{\prime} = W_0 x+ \sum_{i=1}^k G(x)_i \cdot \Delta W_i x.
\end{equation}
Here, $G(x)$ is a $k$-dimensional output of the gating network. A normal way of implementing the gating network is in the following way:
\begin{equation}
    G(x)_i:= exp(x\cdot e_i) / \sum_{j=1}^k exp(x \cdot e_j),
\end{equation}
where $e_i$ denotes the trainable embedding of the expert $\Delta W_i$. However, such a routing mechanism is not suitable for the UML scenario, in which the LoRA pool often changes dynamically. This dynamic nature necessitates continual updates to the Expert Embedding based on modifications to LoRA. Furthermore, newly added LoRAs, which were not included during the initial training phase, are unable to participate in routing due to the lack of prior training.

\subsection{Problem Formulation of Uploadable Machine Learning}
% \item Tasks: $\mathcal{T}=\{T_1, T_2, \cdots, T_n\}$
% \item LoRAs: $\Phi=\{\phi_1, \phi_2, \cdots, \phi_n\}$
% \item mixed tasks: $T_{mix} = \{x, \forall x \in T_1 \lor T_2 \cdots \lor T_n \}$
% \item $y = F(g(\Psi, x), x), x\in T_{mix}$, where $F$ is the corresponding LLM, $g(\cdot)$ represent the LoRA selection process
In this part, we give a formal definition of Uploadable Machine Learning. The UML framework comprises three aspects as shown in Fig.\ref{fig:mix_task}: upstream Edge-side LoRA contributors, the Multi-LoRA serving system on the service side, and downstream heterogeneous requests.

\paragraph{Edge-side Contributors.} Given an original LLM \( L \), suppose we have \( k \) edge-side contributors. Each contributor trains a LoRA module based on its domain dataset, resulting in a set of \( k \) LoRAs, \(\Phi = \{\phi_1, \phi_2, \cdots, \phi_t, \phi_{t+1}^{\prime}, \cdots, \phi_k^{\prime}\}\), where each LoRA \(\phi_i\) is trained on its corresponding task \( T_i \). The LoRAs \(\phi_i^{\prime}\) denote newly uploaded LoRAs that were not visible during the training phase.

\paragraph{Multi-LoRA Serving System.} 
With a LoRA pool $\Phi$ for the LLM $L$, the serving process can be written as: 
\begin{equation}
\label{eq:framework}
    y = F(g(\Phi, x), x, \theta, \gamma),
\end{equation}
where $\theta$ denotes the original parameters of LLM, $g(\Phi, x)$ represents the input-aware LoRA retrieval process, and returns a set of retrieved LoRAs $\Phi_i$. $F(\Phi_i, x_i, \theta)$ depicts the LoRA composition process that integrates the retrieved LoRAs as a plug-in to the original LLM. $\gamma$ denotes the parameters of the gating function that differentiate the contribution of the retrieved LoRAs $\Phi_i$. It is worth noting that during training, the parameters of the retriever $g(\cdot)$ and the gating function $\gamma$ are trained on a set $\mathcal{T}_{train} = \{T_1, \cdots, T_t\}$, leaving the tasks $\{T_{t+1}^{\prime}, \cdots, T_k^{\prime}\}$ unseen for evaluate the generalization for the routing mechanism for unseen LoRAs.

\paragraph{Downstream Heterogeneous Requests.} 
The downstream requests are in mixed task form, where mixed task input can be formulated as $T_{mix} = \{x \mid x \in T_1 \lor T_2 \lor \cdots \lor T_k \}$ where $\lor$ represents the logical disjunction operator. Given the heterogeneous nature of downstream requests, the serving process should provide personalized service for each input prompt.

% Our objective is to develop appropriate mechanisms for LoRA retrieval ($g(\Phi, x)$ in Eq.\ref{eq:framework}) and to establish efficient LoRA composition processes ($F(\Phi_i, x, \theta)$ in Eq.\ref{eq:framework}, where $\Phi_i$ denotes the retrieved LoRAs for $x$).

\begin{figure*}
    \centering
    \includegraphics[width=\linewidth]{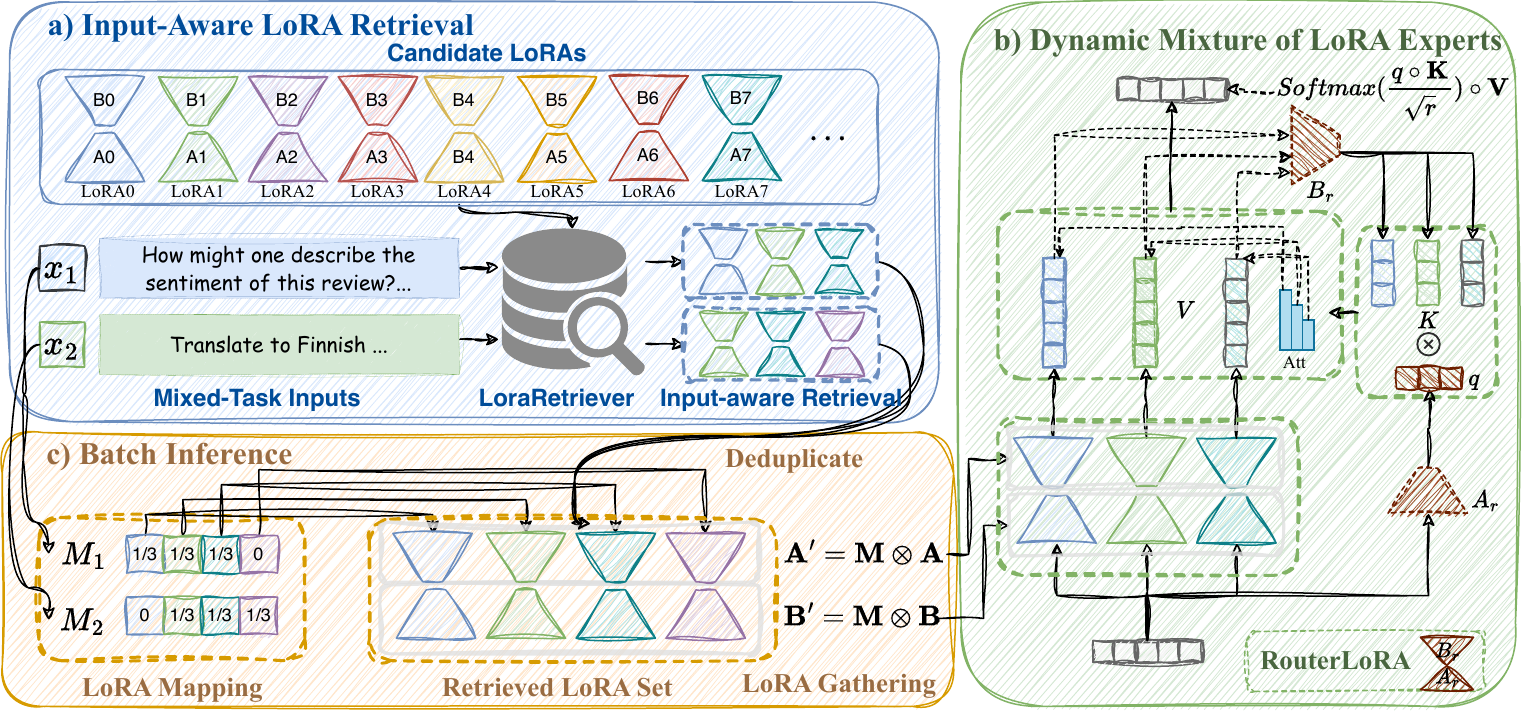}
    \caption{The RAMoLE Framework. This framework, equipped with a pool of candidate LoRAs from various domains/tasks, is designed to offer personalized services tailored to the input provided. It begins by executing an input-aware LoRA retrieval process aimed at identifying LoRAs corresponding to tasks analogous to the input (\S\ref{sec:lora_retrieval}). Subsequently, it employs a specialized LoRA composition mechanism to efficiently utilize the retrieved LoRAs (\S\ref{sec:lora_composition}). By constructing a LoRA mapping matrix for batch inputs, the framework facilitates effective batch inference (\S\ref{sec:batch}).
 }
    \label{fig:framework}
\end{figure*}

\section{Retrieval-Augmented Mixture of LoRA Experts Framework}
In this section, we describe the RAMoLE framework as shown in Fig.\ref{fig:framework} for serving multi-LoRAs in UML. This framework contains three major components: the input-aware LoRA retrieval module (\S\ref{sec:lora_retrieval}), the On-the-fly MoLE module (\S\ref{sec:lora_composition}), and the batch inference strategy (\S\ref{sec:batch}).

\subsection{Input-Aware LoRA Retrieval}
\label{sec:lora_retrieval}
To manage a large pool of LoRAs, the initial stage of our framework involves constructing a LoraRetriever specifically designed to efficiently retrieve the appropriate LoRAs for each input, particularly in scenarios where LoRAs are dynamicaly updated. However, existing approaches fall short of accurately identifying LoRAs under such conditions. MoE-based methods ~\cite{wu2023mole,muqeeth2023soft} struggle to generalize when new LoRAs are introduced due to the fixed selection of LoRAs established during router training. Retrieval methods such as sentence embedding~\cite{reimers2019sentence,ni2021sentence} or task embedding~\cite{achille2019task2vec, zhou-etal-2022-efficiently} fail to map both samples and LoRA into a shared embedding space, limiting their effectiveness in input-aware LoRA retrieval.

%Essentially, we strive for the model to adeptly identify in-domain LoRA (utilized during the retriever's training phase) and to exhibit strong performance in the retrieval of out-of-domain LoRA as well
% To achieve this goal, we propose an instruction-based LoraRetriever ~\cite{su2022one,asai2022task}. Leveraging tailored instructions and fine-tuning processes, this model demonstrates the ability to distinguish between diverse tasks in the embedding space, showcasing robust performance in retrieving out-of-domain LoRA. Specifically, recognizing the considerable differences in training sample distributions across tasks, we facilitate both IID and OOD LoRA retrieval by randomly selecting $m$ training samples for each LoRA and calculating the mean of their instruction embeddings to serve as the LoRA's embedding. This strategy enables effective use of limited data distributions for input-aware retrieval, leading to notable generalization capabilities.

To achieve this goal, we propose to train a retriever via instruction fine-tuning ~\cite{su2022one,asai2022task}, namely LoraRetriever, which can retrieve suitable LoRAs from a massive LoRA pool for a given input sample. The fundamental concept behind LoraRetriever comprises two main steps: (i) First, to embed different task-specific LoRAs into embedding space for facilitating retrieval, we posit that each LoRA can be represented by some data points, which can be obtained by randomly choosing a dozen samples from the training dataset. Then we average their instruction embeddings to represent the embedding of each LoRA. (ii) 
%Through the independent modeling of the LoRA retrieval process, we enable it to adjust to the dynamically updating LoRA pool. Newly updated LoRAs, referred to as out-of-domain (OOD) LoRAs, are thus named because they were unknown during the construction of the retriever.
To improve generalization for unseen LoRAs in LoRA retrieving, we train the retriever through instruction fine-tuning~\cite{su2022one,wei2021finetuned} on a subset of all tasks. Training on a small subset of tasks is designed to simulate scenarios involving the integration of new LoRAs, thereby underscoring our method's generalization abilities via instruction fine-tuning. These two strategies enable the effective use of limited data distributions for input-aware retrieval and can be generalized to unseen LoRAs.

Formally, with a sentence-embedding model $E$, input sequence $x$, and the instruction $I$ for embedding purposes, the instructed embedding can be formulated as $E(I \oplus x)$, where $\oplus$ denotes the concatenation operation. In order to allow the embedding to capture the similarity between different tasks, the instruction is expressed as \textit{"Represent the sentence for similar task retrieval"}. Each LoRA module is embedded with $m$ randomly selected domain-specific samples, expressed as $E(\phi)=\frac{1}{m}\sum^m_{i=1}E(I \oplus x_{i\phi})$. This embedding method integrates both sample-wise and LoRA-module-wise embeddings, facilitating the calculation of similarity between an individual sample and a LoRA module. For measuring the similarity between LoRA module $\phi$ and the input sequence $x$, following ~\cite{ni2021sentence}, we leverage the cosine similarity between the LoraRetriever embeddings: $s(x, \phi, I) = \cos(E(I \oplus x), E(\phi))$.

To improve LoRA retrieval by the retriever and broaden its generalization to unseen LoRAs, we train the embedding model $E$ through instruction fine-tuning on a small subset of tasks. To prevent the need to access new samples, we use previously employed samples for embedding LoRAs as our training data.
Consider $t$ distinct training tasks, represented as $\mathcal{T}_{train} = \{T_1, \cdots, T_t\}$. Following \cite{ni2021sentence}, the training dataset $\mathcal{D}$ comprises paired samples ${(x_i, x_i^+)}$, where each $x_i$ is a sample from a task $T_i \in \mathcal{T}_{train}$, and a positive sample $x_i^+$ is randomly selected from the same task $T_i$. To complement each positive pair, we randomly select $p$ negative pairs ${(x_i, x_{ij}^-)}_{j=1}^p$, ensuring that $x_{ij}^-$ is sourced from tasks outside of $T_i$, thereby $x_{ij}^- \notin T_i$.
The training process is achieved through a contrastive loss~\cite{karpukhin2020dense, izacard2021unsupervised,ni2021sentence} defined as follows:
\begin{equation*}
    \mathcal{L} = \frac{e^{s(x_i,x_i^+, I)/\gamma}}{e^{s(x_i,x_i^+, I)/\gamma} + \sum_{j=1}^p e^{s(x_i,x_{ij}^-, I)/\gamma}},
\end{equation*}
where $\gamma$ is the softmax temperature.

During the LoRA retrieval phase, the top-$k$ LoRAs are retrieved according to their similarity to the input $x$. This process can be formulated as follows:
\begin{equation*}
    g(x_i, \Phi) :=  \Phi_i = \mathrm{TopK} \{s(\phi_j,x_i,I) , \phi_j \in \Phi \}.
    % g(x_i, \Phi) := \Phi_i = \underset{\phi_j \in \Phi_i, \Phi_i \subset \Phi,|\Phi_i| = k}{\mathrm{argmax}} \ s(\phi_j, x_i, I).
\end{equation*}

\subsection{On-the-fly Mixture of LoRA Experts}
\label{sec:lora_composition}
After retrieving the top-k LoRAs, $\Phi_i$, for an input $x_i$, we proceed to integrate these LoRAs into the LLM with parameter $\theta$. This section delves into the proposed on-the-fly Mixture of LoRA Experts (DMoLE) mechanism for LoRA composition. We start by introducing the vanilla LoRA composition strategies as described in \cite{zhao2024loraretriever}, and illustrate the limitations of these vanilla methods. Subsequently, we offer a formal description of the proposed on-the-fly MoLE mechanism, which incorporates an additional LoRA serving as a router, namely RouterLoRA. Its primary purpose is to learn to route among retrieved LoRAs, effectively using cross-attention to assign weights to each one and decouple the routing process from specific LoRA modules.

\subsubsection{Vanilla LoRA Composition}
The vanilla LoRA composition comprises two training-free methods: the Mixture of LoRAs and the Fusion of LoRAs.
\paragraph{Mixture of LoRAs}
The mixture of LoRAs strategy involves the aggregation of the outputs of each submodule within the assembled LoRAs. Let us denote $\mathbf{A} = \{A_1, A_2, \ldots, A_n\}$ and $\mathbf{B} = \{B_1, B_2, \ldots, B_n\}$ as the sets representing submodules within $n$ LoRAs. For an input $x_i$, the output derived from the mixture of LoRAs can be expressed as:
\begin{equation}
x_i^{\prime} = \frac{1}{n} \sum_{j=1}^n B_j A_j x_i,
\end{equation}
where $x_i^{\prime}$ denotes the output. This process signifies the integration of each LoRA module's output, effectively blending their contributions to form a unified output.

\paragraph{Fusion of LoRAs}
In contrast to the Mixture method, which combines the output of different LoRAs, fusing the parameters of these LoRAs presents an alternative composition strategy. Let the parameters of each LoRA $\phi_i$ be denoted by $\Theta_i$. The parameter of the fused LoRA is then represented as $\Theta_{\text{fusion}} = \frac{1}{k} \sum_{j=1}^k \Theta_j$. This formulation allows the fused parameter to function akin to a single LoRA.

\paragraph{Limitation of Vanilla Composition}
Since the LoRAs for different tasks are trained independently, the model is significantly affected by data heterogeneity during the parameter fusion process. This results in catastrophic collapses in performance on respective tasks after fusion due to parameter interference, a phenomenon that becomes more pronounced with the increasing number of composed LoRAs~\cite{zhao2024loraretriever}. Consequently, the Fusion of LoRAs method is not suitable for application in scenarios involving UML. On the other hand, the mixture of LoRAs method, which averages the outputs of different LoRAs, achieves better performance, especially on unknown tasks. However, because it uniformly assigns the same weight to the output of each LoRA without differentiation, this method fails to fully leverage the most optimal LoRA for the task, leading to a suboptimal model performance.

\subsubsection{On-the-fly Mixture of LoRA Experts with RouterLoRA}
To allocate weights for the retrieved top-k LoRAs $\Phi_i$ in layer-level routing, the most straightforward idea is to train an additional router using the MoLE method to assign weights to each LoRA. However, since the traditional MoLE method requires fixing the selection of LoRAs during training and ensuring consistency between the LoRAs loaded during training and inference, it is unable to adapt to the dynamically changing LoRA pool in UML scenarios, making it unsuitable for such scenarios. 
% To address this issue, we propose a novel MoE mechanism that trains an extra LoRA to learning as a router, using an attention mechanism to assign weights to the retrieved LoRAs.
To address this challenge, we propose a novel MoLE mechanism that includes an additional LoRA module specifically designed for \textit{learning to route}. This module leverages an attention mechanism to dynamically assign weights to different LoRAs, optimizing the routing process.
In this way, we decouple the router from the LoRA module for routing, achieving a more dynamic MoLE mechanism that can effectively generalize to newly uploaded LoRAs.

% Formally, consider $\mathbf{A} = {A_1, A_2, \ldots, A_k}$ and $\mathbf{B} = {B_1, B_2, \ldots, B_k}$ as the sets of top-$k$ LoRAs retrieved for a layer. A RouterLoRA, equipped with parameters $A_r$ and $B_r$, is integrated into the LLM. It functions as a router to determine the appropriate weights for each plugged-in LoRA module. The entire routing process is conducted in a manner similar to the cross-attention mechanism [tocite]. Firstly, for the input $x$, it passes through each LoRA module ($A_i$, $B_i$) to obtain $v_i = B_i A_i x$. And the query $q$ is obtained from $q = A_r x$, where $q \in \mathbb{R}^{b\times l \times r}$. The key of i-th LoRA module can be obtained from $k_i = B_r^T v_i$.

Formally, consider $\mathbf{A} = \{A_1, A_2, \ldots, A_n\}$ and $\mathbf{B} = \{B_1, B_2, \ldots, B_n\}$ as the sets of top-$n$ LoRAs retrieved for a layer. A RouterLoRA, with parameters $A_r$ and $B_r$, is integrated into the LLM. It functions as a router to determine the appropriate weights for each plugged-in LoRA module.
The entire routing process operates through a mechanism akin to cross-attention operation. Initially, for an input $x$, it is processed by each LoRA module $(A_i, B_i)$ to generate
\begin{equation}
    v_i = B_i A_i x,
\end{equation}
which can be seen as a transformation of $x$ through a module-specific linear transformation followed by another module-specific transformation.
Furthermore, the query vector $q$ is derived from the router's parameters using:
\begin{equation}
    q = A_r x,
\end{equation}
where $q \in \mathbb{R}^{b\times l \times r}$ (assuming $b$ is the batch size, $l$ is the sequence length, and $r$ is the dimensionality of the query vector). Each $i$-th LoRA module contributes a key vector $k_i$, calculated as:
\begin{equation}
    k_i = B_r^T v_i,
\end{equation} 
where $k \in \mathbb{R}^{b\times l \times r}$.
This transformation aims to map the output $v_i$ into a space where it can be compared against the query $q$.

The next step involves calculating the attention weights. Each $k_i$ is compared to $q$ using a dot product, resulting in a score that measures the alignment or relevance of each LoRA output to the query. These scores are then normalized through a softmax function to form attention weights:
\begin{eqnarray}
    &s_i = \langle q, k_i \rangle / \sqrt{r} \\
    &\alpha = \text{Softmax}(s_1, s_2, \ldots, s_n).
\end{eqnarray}

These attention weights $\alpha_i$ dictate the importance of each $v_i$ in the final representation. The weighted sum of all $v_i$, modulated by $\alpha_i$, produces the output of the RouterLoRA:
\begin{equation}
    x^{\prime} = \sum_{i=1}^k \alpha_i v_i.
\end{equation}

This output $x^{\prime}$ then serves as the adjusted representation for the original input $x$ within the context of the LLM, reflecting a dynamically weighted integration of different transformations offered by the LoRA modules. Thus, RouterLoRA employs an attention mechanism to adaptively route and differentiate LoRA modules across different layers, resulting in more fine-grained routing.

\paragraph{Training of RouterLoRA} Similar to the training process of LoraRetriever, here we leverage $t$ distinct training tasks, represented as $\mathcal{T}_{train} = \{T_1, \cdots, T_t\}$, and the training set $\mathcal{D}_{train} = \{(x,y) \in \mathcal{T}_{train}\}$. To test the generalization of the Dynamic MoLE mechanism, we train on only 40\% of the LoRAs in the LoRA pool and perform zero-shot routing on the remaining LoRAs. And the training set of LoRAs can be denoted as $\Phi_{train} = \{\phi_1, \cdots, \phi_t\}$. Besides, we denote the parameters of the RouterLoRA as $\gamma$. The training loss can be formulated in the following way:
\begin{equation}
    \mathcal{L}_{train} = \frac{1}{|\mathcal{D}_{train}|} \sum_{x, y \in \mathcal{D}_{train}} \mathcal{L}(f(x; \Phi_{train}, \gamma, \theta), y).
\end{equation}
In this formulation, only $\gamma$ is trainable, focusing the training process on optimizing the router's parameters while keeping the parameters of individual LoRAs fixed. Through this targeted training, we successfully developed RouterLoRA to act as a router, capable of distinguishing between LoRA experts based on different hidden states, thus facilitating a more flexible and dynamic routing mechanism.

\paragraph{Random Dropout LoRA modules for Improving Generalization.} Loading all LoRAs directly into the model and training RouterLoRA on specific tasks significantly improves performance in in-distribution (IID) scenarios by efficiently identifying and routing to the optimal LoRA module. However, this method is less effective in out-of-distribution (OOD) scenarios, where no single LoRA perfectly fits the inference needs of downstream tasks. In these cases, combining different LoRAs is crucial for enhancing zero-shot generalization. Direct training of RouterLoRA often overlooks the complexities of unseen tasks, focusing mainly on optimizing performance within familiar IID tasks and neglecting the variability of OOD scenarios. Consequently, the model lacks the adaptability and robustness needed for effective generalization in new, unanticipated contexts, leading to suboptimal zero-shot performance. To mitigate this issue, we propose a strategy of randomly dropout LoRA modules during the training process to enhance our method's performance in OOD scenarios. Formally, we redefine the training objective to incorporate module-level dropout as follows:

\begin{equation}
    \mathcal{L}_{train} = \frac{1}{|\mathcal{D}_{train}|} \sum_{x, y \in \mathcal{D}_{train}} \mathcal{L}(f(x; Dropout(\Phi_{train},p), \gamma, \theta), y),
\end{equation}
% \begin{equation}
% \scalebox{0.95}{$
%     \mathcal{L}_{train} = \frac{1}{|\mathcal{D}_{train}|} \sum_{x, y \in \mathcal{D}_{train}} \mathcal{L}(f(x; \text{Dropout}(\Phi_{train},p), \gamma, \theta), y)
% $}
% \end{equation}
where $p$ denotes the dropout rate, here, the function $Dropout(\Phi_{train},p)$ selectively deactivates a subset of modules $\Phi_{train}$ at a rate $p$, enabling us to train a more robust model by introducing variability and reducing overfitting.

% \subsubsection{Fusion of LoRAs}
% In contrast to the Mixture method, which combines the output of different LoRAs, fusing the parameters of these LoRAs presents an alternative composition strategy.

% Let the parameters of each LoRA $\phi_i$ be denoted by $\Theta_i$. The parameter of the fused LoRA is then represented as $\Theta_{\text{fusion}} = \frac{1}{k} \sum_{j=1}^k \Theta_j$. This formulation allows the fused parameter to function akin to a single LoRA.

\subsection{Batch Inference of Multiple LoRAs}
\label{sec:batch}
Implementing batch inference in the presence of multiple LoRAs and diverse composition diagrams poses a significant technical challenge. To address this, we introduce a unique approach for batch inference. Our method involves processing a batch of samples denoted as $X \in \mathbb{R}^{b \times l \times d}$, where $b$, $l$, and $d$ denote the batch size, sequence length, and sample dimensionality, respectively. 
% We retrieve $k$ LoRAs for each individual sample $x_i$ and construct a collective set of LoRAs, represented as $\Phi_\mathcal{B}$. We maintain the distinctness of the set $\Phi_\mathcal{B}$ by removing any redundant LoRAs, considering the potential for overlapping LoRAs in different samples, where the set contains $p$ LoRAs and $p \leq bk$. A mapping vector $M_i$ is then created for each sample $x_i$, indicating the indices of its corresponding LoRAs in the set $\Phi_\mathcal{B}$.
For each input $x_i$ and its retrieved LoRAs $\Phi_i$ within the same batch, we aggregate these LoRAs into a collective set denoted by $\Phi_\mathcal{B}$. To ensure the uniqueness of $\Phi_\mathcal{B}$, we eliminate duplicates, mindful of the possibility that retrieved LoRAs may overlap across different samples. The resulting set $\Phi_\mathcal{B}$ comprises $p$ unique LoRAs, where $p \leq bk$. For every sample $x_i$, a $p$ dimension mapping vector $M_i$ is generated, which specifies the indices of its corresponding LoRAs within $\Phi_\mathcal{B}$.

The LoRA mapping vectors are combined into a matrix $\mathbf{M} \in \mathbb{R}^{b\times p}$. The parameters of a submodule in LoRA can be denoted as $A$ and $B$, and are concatenated within the batched LoRAs $\Phi_{\mathcal{B}}$ to obtain $\mathbf{A} \in \mathbb{R}^{p\times r\times d}$ and $\mathbf{B} \in \mathbb{R}^{p\times d\times r}$. We first gather the LoRA modules for each input within the batch through $\mathbf{A}^\prime =\mathbf{M} \otimes \mathbf{A}$ and $\mathbf{B}^\prime =\mathbf{M} \otimes \mathbf{B}$ where $\otimes$ denotes the gather operation. By gathering the corresponding parameters for each input, we obtain the matrix $\mathbf{A}^\prime \in \mathbb{R}^{b\times k \times r \times d}$ and $\mathbf{B}^\prime \in \mathbb{R}^{b\times k \times d \times r}$. Then we can obtain the value matrix $\mathbf{V}$ through $\mathbf{V} = \mathbf{B}^\prime \circ \mathbf{A}^\prime \circ X$, where $\mathbf{V} \in \mathbb{R}^{k\times b \times l \times d}$ and we extend the symbol $\circ$ to denote potential broadcasting as~\cite{wen2023batched}. Then the key matrix is computed through the RouterLoRA in the following way:
\begin{eqnarray}
    & q = A_r \circ X \\
    & \mathbf{K} = B_R^T \circ \mathbf{V}
\end{eqnarray}
where $q\in \mathbb{R}^{1\times b\times l\times r}$ and $\mathbf{K} \in \mathbb{R}^{k\times b\times l\times r}$. The output is computed by $X^\prime = Softmax(\frac{q \circ \mathbf{K}}{\sqrt{r}}) \circ \mathbf{V}$.
In this way, we successfully perform personalized inference for each input sample, allowing efficient batch inference.

% The LoRA mapping vectors are combined into a matrix $\mathbf{M} \in \mathbb{R}^{b\times p}$. The parameters of a submodule in LoRA can be denoted as $A$ and $B$, and are concatenated within the batched LoRAs $\Phi_{\mathcal{B}}$ to obtain $\mathbf{A} \in \mathbb{R}^{p\times r\times d}$ and $\mathbf{B} \in \mathbb{R}^{p\times d\times r}$.
% The batch inference process of the mixture of LoRAs can be formulated as follows:
% \begin{equation}
%     X^{\prime} = \mathbf{M}\circ (\mathbf{B} \circ\mathbf{A} \circ X),
% \end{equation}
% where we denote the batched output of a layer of multiple LoRA as $X^{\prime}\in \mathbb{R}^{b \times l \times d}$ and extend the symbol $\circ$ to denote potential broadcasting as~\cite{wen2023batched}.

% The batch inference process of LoRA fusion can be formulated as
% \begin{equation}
%     X^{\prime} = (\mathbf{M} \circ \mathbf{B})(\mathbf{M} \circ\mathbf{A}) \circ X.
% \end{equation}
% selected_lora_A = torch.einsum('ij,jkl->ikl', self.lora_attention, self.lora_A_matrix)
% selected_lora_B = torch.einsum('ij,jkl->ikl', self.lora_attention, self.lora_B_matrix)
% # intermediate_result = torch.einsum('aec,abc->aeb', dropout(x), selected_lora_A)/(self.lora_num)
% # result += torch.einsum('aeb,acb->aec', intermediate_result, selected_lora_B)*2/(self.lora_num)
% intermediate_result = torch.einsum('aec,abc->aeb', dropout(x), selected_lora_A)
% result += torch.einsum('aeb,acb->aec', intermediate_result, selected_lora_B)*2

\section{Experiments}
This section outlines the evaluation framework for assessing different approaches in our Uploadable Machine Learning setting. We conducted a detailed comparison of the performance differences of various LoRA composition strategies in different settings, to demonstrate the superiority of the proposed DAMoLE method. Furthermore, a comprehensive analysis of the proposed RAMoLE framework is presented.

\subsubsection{Training And Evaluation Datasets}
We leverage a subset of flan-v2 datasets ~\cite{wei2021finetuned} as shown in Fig.\ref{fig:tasks} for LoRA expert training and mixed-task dataset generation. We adopt the task cluster as described by ~\cite{wei2021finetuned}. The details of datasets can be found in the Appendix. \ref{sec:eval_dataset}.

\subsubsection{Base Model \& LoRA Configuration.}
To test various methods in the mixed-task scenarios, we leverage Llama-2-\{7b,13b\}~\cite{touvron2023llama} as base models and train a range of LoRAs for a spectrum of tasks. We selected a portion of the Flan-v2 datasets~\cite{wei2021finetuned} to train 48 LoRAs for a spectrum of tasks covering Natural Language Understanding (NLU) and Natural Language Generation (NLG). Following the categorization by \cite{wei2021finetuned}, these tasks can be grouped into 10 distinct task clusters. We train each LoRA according to the Alpaca~\cite{alpaca} format and rank $r$, and the scaling hyperparameter $\alpha$ are set to 6 and 12, respectively. 

\begin{table*}
\caption{The average performance of each task cluster. 
% The full results of each task are shown in Tab.\ref{tab:full_results} \& \ref{tab:full_results_13b}. 
"IID" means that RAMoLE can access any LoRA for every test sample, encompassing the LoRA specific to the sample's task. "OOD" indicates that for each test sample, we mask the LoRA associated with its specific task during the retrieval phase. Consequently, no sample can access its ideal LoRA, allowing us to assess the RAMoLE's cross-task generalization capability. The performance of perfectly selected corresponding LoRA for each sample is colored in gray. We have bolded the best performance of each task and underlined the best performance in the "OOD" setting.}
\centering
\resizebox{\linewidth}{!}{  
\begin{tabular}{l c c c c c c c c c c c c c c c} 
\toprule
\multirow{2}{*}{Task} &
\multirow{2}{*}{\begin{tabular}[c]{@{}c@{}}Perfect\\Selection\end{tabular}} &
\multicolumn{2}{c}{RAMoLE} & 
\multicolumn{2}{c}{Selection} & 
\multicolumn{2}{c}{Fusion} & 
\multicolumn{2}{c}{Mixture} & 
\multirow{2}{*}{\begin{tabular}[c]{@{}c@{}}MoE\\Top1\end{tabular}} &
\multirow{2}{*}{\begin{tabular}[c]{@{}c@{}}MoE\\Top3\end{tabular}} &
\multirow{2}{*}{\begin{tabular}[c]{@{}c@{}}MoE\\Soft\end{tabular}} &
\multirow{2}{*}{\begin{tabular}[c]{@{}c@{}}SME-\\AR\end{tabular}} &
\multirow{2}{*}{\begin{tabular}[c]{@{}c@{}}Adapter\\Soup\end{tabular}} &
\multirow{2}{*}{\begin{tabular}[c]{@{}c@{}}LoRA\\Hub\end{tabular}}\\
\cline{3-10}
& & IID & OOD & IID & OOD & IID & OOD & IID & OOD & & & & \\
\midrule
\multicolumn{16}{c}{\textit{w/ Llama2-7b}} \\\hline
$\text{Struct to Text}_{Rouge-1}$ & \cellcolor[gray]{0.8} 59.1 & \textbf{58.4} & \underline{50.0} & 56.8 & 45.2 & 44.5 & 41.0 & 51.2 & 45.3 & 41.3 & 41.8 & 43.2 & 43.3 & 3.5 & 31.9 \\
$\text{Struct to Text}_{Rouge-2}$  & \cellcolor[gray]{0.8} 36.1 & \textbf{32.6} & \underline{26.1} & \textbf{33.6} & 23.2 & 22.6 & 20.2 & 26.3 & 22.9 & 19.3 & 19.9 & 20.7 & 21.3 & 0.9 & 15.1 \\
$\text{Struct to Text}_{Rouge-l}$  & \cellcolor[gray]{0.8} 48.6 & \textbf{51.9} & \underline{44.6} & 46.4 & 35.3 & 34.5 & 31.7 & 41.0 & \underline{35.5} & 32.6 & 32.8 & 33.8 & 33.9 & 3.3 & 24.9 \\
$\text{Translation}_{BLEU}$ & \cellcolor[gray]{0.8}  13.1 & \textbf{12.8} & 11.9 & \textbf{12.8} & 12.0 & 12.2 & \underline{12.3} & \textbf{12.8} & 12.2 & 9.5 & 10.5 & 10.7 & 11.0 & 1.4 & 8.5 \\
\midrule
COMMONSENSE & \cellcolor[gray]{0.8} 62.5 & \textbf{58.5} & 55.0 & 55.5 & 46.0 & 51.0 & 48.0 & \textbf{61.5} & 50.0 & 54.5 & 52.0 & 51.5 & 50.0 & 46.0 & 17.5 \\
SENTIMENT & \cellcolor[gray]{0.8} 90.0 & \textbf{90.5} & 92.0 & 89.5 & 89.0 & 79.0 & 78.5 & 89.5 & 90.5 & 70.0 & 75.0 & 74.5 & 74.0 & 73.5 & 0.5 \\
READING Comp. & \cellcolor[gray]{0.8} 67.3 & \textbf{52.0} & 47.7 & 51.7 & 40.3 & 47.3 & 45.0 & 51.3 & 47.3 & 48.7 & 47.7 & 48.7 & 45.7 & 40.7 & 2.7 \\
CLOSE-BOOK QA & \cellcolor[gray]{0.8} 45.0 & \textbf{46.0} & 45.5 & 40.0 & 43.0 & 41.0 & 37.5 & 45.0 & \underline{48.5} & 40.5 & 38.5 & 40.0 & 32.0 & 31.5 & 1.0 \\
COREFERENCE & \cellcolor[gray]{0.8} 52.0 & 61.0 & \underline{52.0} & 50.0 & 46.0 & 47.0 & \underline{53.0} & \textbf{63.0} & 49.0 & 61.0 & 59.0 & 57.0 & 58.0 & 43.0 & 1.0 \\
READ. COOMP. W/ COM & \cellcolor[gray]{0.8} 69.0 & 64.0 & \underline{42.0} & \textbf{69.0} & 30.0 & 35.0 & 19.0 & 46.0 & 40.0 & 31.0 & 29.0 & 29.0 & 23.0 & 14.0 & 3.0 \\
PARAPHRASE & \cellcolor[gray]{0.8} 65.5 & 57.0 & \underline{46.5} & \textbf{58.0} & 45.5 & 45.5 & 44.0 & 56.5 & 45.5 & 42.0 & 38.5 & 36.0 & 34.5 & 46.5 & 1.0 \\
NLI & \cellcolor[gray]{0.8} 72.3 & 65.9 & 66.2 & \textbf{70.0} & 60.6 & 51.4 & 53.8 & 67.9 & \underline{64.3} & 50.3 & 49.6 & 48.3 & 50.8 & 62.4 & 10.5 \\
Overall & \cellcolor[gray]{0.8} 55.4 & \textbf{51.8} & \underline{47.7} & 51.2 & 43.0 & 41.6 & 40.2 & 49.8 & 45.6 & 40.3 & 39.9 & 39.8 & 39.1 & 32.4 & 10.1 \\
\midrule
\multicolumn{16}{c}{\textit{w/ Llama2-13b}} \\\hline
$\text{Struct to Text}_{Rouge-1}$ & \cellcolor[gray]{0.8} 65.4 & 60.7 & 49.6 & \textbf{62.6} & 49.4 & 52.7 & 49.7 & 57.7 & 52.1 & 46.8 & 47.0 & 48.5 & 48.3 & 7.1 & 39.3 \\
$\text{Struct to Text}_{Rouge-2}$  & \cellcolor[gray]{0.8} 40.8 & 35.1 & 26.7 & \textbf{38.2} & 25.8 & 29.2 & 26.8 & 32.6 & 28.1 & 24.5 & 25.1 & 25.7 & 25.2 & 2.5 & 20.7 \\
$\text{Struct to Text}_{Rouge-l}$  & \cellcolor[gray]{0.8} 58.7 & 53.8 & 42.6 & \textbf{56.0} & 42.9 & 45.9 & 43.2 & 50.8 & 45.4 & 41.1 & 41.9 & 42.7 & 42.2 & 6.4 & 34.6 \\
$\text{Translation}_{BLEU}$ & \cellcolor[gray]{0.8} 12.9 & 14.0 & 14.0 & 12.9 & 12.7 & \textbf{14.6} & 14.1 & \textbf{14.6} & 14.1 & 11.8 & 12.4 & 11.9 & 12.4 & 0.8 & 10.2 \\
\midrule
COMMONSENSE & \cellcolor[gray]{0.8} 69.5 & \textbf{69.5} & 68.0 & 59.0 & 47.5 & 61.0 & 56.0 & 64.0 & 60.5 & 65.0 & 66.0 & 64.0 & 61.0 & 17.5 & 34.0 \\
SENTIMENT & \cellcolor[gray]{0.8} 90.0 & \textbf{91.5} & \textbf{91.5} & 90.5 & 91.0 & 87.0 & 83.5 & \textbf{91.5} & \underline{91.5} & 90.0 & 89.5 & 90.0 & 89.0 & 79.5 & 11.0 \\
READING Comp. & \cellcolor[gray]{0.8} 76.0 & \textbf{62.3} & \underline{54.3} & 60.3 & 48.0 & 56.7 & 49.3 & 60.3 & 51.3 & 53.7 & 53.3 & 52.3 & 51.3 & 48.7 & 3.3 \\
CLOSE-BOOK QA & \cellcolor[gray]{0.8} 64.0 & 61.0 & 56.0 & 60.0 & 53.0 & 62.0 & 58.0 & \textbf{63.0} & \underline{61.0} & 59.5 & 57.5 & 58.5 & 57.5 & 34.5 & 6.5 \\
COREFERENCE & \cellcolor[gray]{0.8} 74.0 & \textbf{76.0} & 60.0 & 75.0 & 65.0 & 55.0 & 59.0 & \textbf{76.0} & \underline{64.0} & 61.0 & 62.0 & 56.0 & 57.0 & 55.0 & 10.0 \\
READ. COOMP. W/ COM & \cellcolor[gray]{0.8} 82.0 & 78.0 & 57.0 & \textbf{80.0} & 33.0 & 57.0 & 49.0 & 78.0 & \underline{58.0} & 51.0 & 48.0 & 49.0 & 49.0 & 13.0 & 14.0 \\
PARAPHRASE & \cellcolor[gray]{0.8} 77.5 & 67.0 & \underline{60.5} & \textbf{68.0} & 52.5 & 55.5 & 45.5 & 71.0 & 55.5 & 50.0 & 52.5 & 47.5 & 52.0 & 64.0 & 2.5 \\
NLI & \cellcolor[gray]{0.8}  82.4 & 78.8 & 75.3 & \textbf{78.9} & 70.2 & 69.8 & 66.4 & 78.1 & \underline{75.7} & 67.7 & 71.0 & 67.4 & 66.6 & 67.5 & 14.9 \\
Overall & \cellcolor[gray]{0.8} 63.6 & \textbf{59.6} & \underline{53.7} & 59.0 & 48.9 & 52.7 & 48.9 & 58.8 & 53.5 & 50.4 & 51.2 & 50.0 & 49.7 & 34.9 & 16.0 \\
\bottomrule
\end{tabular}
}
\vspace{-0.15in}
\label{tab:main_sum}
\end{table*}

\subsubsection{Evaluation Dataset.} For constructing the evaluation dataset, we randomly chose 50 samples from the test set for each task used in training 48 LoRAs, subsequently mixing and shuffling these samples to form a unified dataset with 2400 data entries.

\subsubsection{Baseline Methods.} 
We compared our method with the following baselines: 
(1) \textbf{Mixture of Experts} ~\cite{zhu2023sira,zadouri2023pushing,liu2023moelora,wang2022adamix,wu2023mole}. Many works have considered coordinating different adapters through MoE, and here we explored three distinct variants: one employing a soft mixture of experts and the other utilizing discrete routing (top1 and top3).It is worth noting that in this paper, our primary focus is on post-hoc MoE training. This implies that the LoRA experts have been previously trained, and only the parameters of the router are trainable during the training of the MoE methods.
(2) \textbf{SMEAR} ~\cite{muqeeth2023soft} introduces the concept of adaptive routing by performing a weighted average of different adapters' parameters to utilize various experts effectively. For the MoE and SMEAR baselines, challenges arise in scaling due to training confined to a limited set of LoRAs. Consequently, we strategically selected a dedicated LoRA expert for each domain to specialize in router training.
(3) \textbf{AdapterSoup} ~\cite{chronopoulou2023adaptersoup} uniformly selects the corresponding LoRAs for the entire downstream task, which lacks the ability to provide personalized service for diverse requests.
(4) \textbf{LoRAHub} ~\cite{huang2023LoRAHub} enables black-box optimization to learn the weights of various LoRA parameters, thereby facilitating weighted parameter averaging for specific downstream tasks. In our implementation, we conformed to the default setting, which entails randomly selecting 20 LoRAs from the available LoRA pool and performing weighted parameter averaging.
For the MoE, SMEAR, and LoRAHub approaches, we selected 20 data samples from the training datasets of all tasks to serve as their training data.

\begin{table}
\caption{Comparison of Sentence Embedding Techniques in LoRA Retrieval: The notation LoraRetriever\textsuperscript{k\%} signifies that the model underwent supplementary training on $k$ percent of the tasks. The performance of the selected retriever model in the evaluation phase is highlighted in gray.}
\centering
\resizebox{\linewidth}{!}{  
\begin{tabular}{lcccc}
\toprule
\textbf{Method} & \textbf{Top 1} & \textbf{Top 3} & \textbf{Top 5} & \textbf{Top 8} \\ 
\midrule
all-mpnet-base-v2 & 58.40 & 78.26 & 84.77 & 90.24 \\
all-MiniLM-L6-v2 & 51.73 & 73.11 & 80.54 & 87.18 \\
msmarco-distilbert-cos-v5 & 45.84 & 66.01 & 75.14 & 82.67 \\
% sentence-t5-large & 63.52 & 86.37 & 92.07 & 96.44 \\
gtr-t5-xl & 53.19 & 69.72 & 77.41 & 83.59 \\
\midrule
LoraRetriever \textsuperscript{0\%} & 60.80 & 79.29 & 85.57 & 91.58 \\
 \cellcolor[gray]{0.8}
 LoraRetriever \textsuperscript{40\%} & \cellcolor[gray]{0.8} 63.16 & \cellcolor[gray]{0.8} 89.09 & \cellcolor[gray]{0.8} 95.45 & \cellcolor[gray]{0.8} 98.97 \\
% INSTRUCTOR \textsuperscript{60\%} & 65.12 & 91.89 & 96.97 & 99.46 \\
LoraRetriever \textsuperscript{100\%} & \textbf{74.08} & \textbf{97.37} & \textbf{99.15} & \textbf{99.82} \\
\bottomrule
\end{tabular}
}
\label{tab:retrival_acc}
\end{table}

\subsubsection{Implementation of Baseline Methods.} 
\textbf{MoE baselines.}
We use $E$ to denote the LoRA expert and $R$ to denote the router. The MoE methods can be expressed in the following way:
\begin{equation}
    y = \sum_{i=i}^k R(x)_i E_i(x).
\end{equation}
We implied two variants of the MoE routing mechanism. (1) \textbf{Dense Gating.} Following ~\cite{zadouri2023pushing}, the router network consists of a dense layer with trainable parameter $W_g$, and the gating score could be obtained through a softmax function by:
\begin{equation}
    s_i = R(x)_i = softmax(W_g^T x),
\end{equation}

(2) \textbf{Sparse Gate}. To maintain the sparsity while training, we leverage the Gumbel softmax trick as ~\cite{muqeeth2023soft, nie2021dense}, where the router can be written as:
\begin{equation}
    \hat{R}(x)_i = \frac{(log(R(x)_i)+g_i)/ \tau}{\sum_{i=1}^k exp((log(R(x)_i)+g_i)/\tau)}
\end{equation}
where $g_i \sim \text{Gumbel}(0,1)$ and $\tau$ is the temperature. 

Due to MoE not being easily scalable and arbitrarily adding new LORAs, we randomly selected a LoRA as an expert for each task cluster in the experiment and trained the corresponding Router's parameters. We randomly selected 20 samples for each task during training to form a unified dataset for parameter training.

\textbf{SMEAR}
SMEAR ~\cite{muqeeth2023soft} does not perform routing aggregation on the Adapter output but rather aggregates the Adapter at the parameter level. We adopt the same setting as the MoE methods, and the results could be calculated in the following way:
\begin{equation}
    \Theta_{SMEAR} = \sum_{i=i}^k R(x)_i \Theta_i,
\end{equation}
where $\Theta_i$ denote the parameter of the LoRA-$i$. 

\textbf{AdapterSoup}
AdapterSoup ~\cite{chronopoulou2023adaptersoup}, for new downstream tasks, retrieves the parameters that need to be involved in aggregation through sentence bert and performs weight-space averaging on these parameters to adapt to the new domain. We have uniformly retrieved 3 LoRAs for mixed-task to test their capabilities under mixed-task conditions.

\textbf{LoRAHub}
LoRAHub ~\cite{huang2023LoRAHub} also aggregates 20 LoRAs randomly for new downstream tasks. In order to learn the weight of LoRA, a black-box optimization method is employed to learn the weight of each LoRA without calculating the gradients of the large model. It performs weighted averaging at the parameter level. Similar to the training process of MoE, we randomly selected 20 samples for each task to form a unified training dataset for black-box optimization.

\subsubsection{Implementation of RAMoLE.} 
To train RAMoLE, we divide our training process into two stages: first, training the LoraRetriever, and second, training the RouterLoRA.
To train the LoraRetriever, we continue to perform instruction fine-tuning based on Instructor-xl ~\cite{su2022one}. The training data consisted of only 40\% of the tasks used to train task-specific LoRAs, with each task represented by 20 samples randomly selected from its respective LoRA training set.
% , showcasing the generalization of utilizing a retriever as a routing mechanism
In this process, we categorized samples from the same LoRA as positive examples, while those from different LoRAs were considered negative examples. 

For training the RouterLoRA, we also use only 40\% of the tasks and the corresponding LoRAs for training, leaving the remaining LoRAs unseen to test the generalization of the on-the-fly MoE mechanism. During the training process, we load all these LoRAs into the LLM and randomly dropout 50\% of the LoRAs to enhance the robustness of the router training. The RouterLoRA shares the same configuration as the task LoRA, with a rank of $r=6$ and a scaling hyperparameter of $\alpha=12$.

Additionally, we compared three vanilla LoRA composition strategies: (1) \textbf{Selection}, which involves selecting the highest-ranked (top-1) retrieved LoRA and applying it singularly, serving as a variant of the Mixture and Fusion methods; (2) \textbf{Mixture}, which averages the outputs of each submodule from the top-$k$ retrieved LoRAs; and (3) \textbf{Fusion}, a method that averages the parameters of the top-$k$ retrieved LoRAs. Throughout our experiments, $k=3$ was established as the default setting.

% \input{tables/main_res_gen.tex}

% \begin{table}
% \centering
% \begin{tabular}{lccc}
% \toprule
% Methods & 0\% & 40\% ($\Delta$\%) & 100\% ($\Delta$\%) \\ \midrule
% Selection & 57.99 & 62.42 (+7.64\%) & 64.01 (+2.55\%) \\
% Fusion & 51.50 & 51.50 (+0.00\%) & 52.27 (+1.49\%) \\
% Mixture & 62.24 & 63.54 (+2.09\%) & 64.19 (+1.02\%) \\
% \bottomrule
% \end{tabular}
% \caption{Average Performance of LoraRetriever on NLU Tasks Across Different LoRA Composition Strategies and Task Training Percentages.}
% \label{table:methods_improvement}
% \end{table}

\subsubsection{Metrics.}
Following \cite{wei2021finetuned}, we assess the performance on the "Struct to Text" task using Rouge-\{1, 2, L\} and on the "Translation" tasks using BLEU. Additionally, for the NLU tasks, we evaluate the exact match accuracy of each method.

\begin{figure*}
    \centering
    \includegraphics[width=\linewidth]{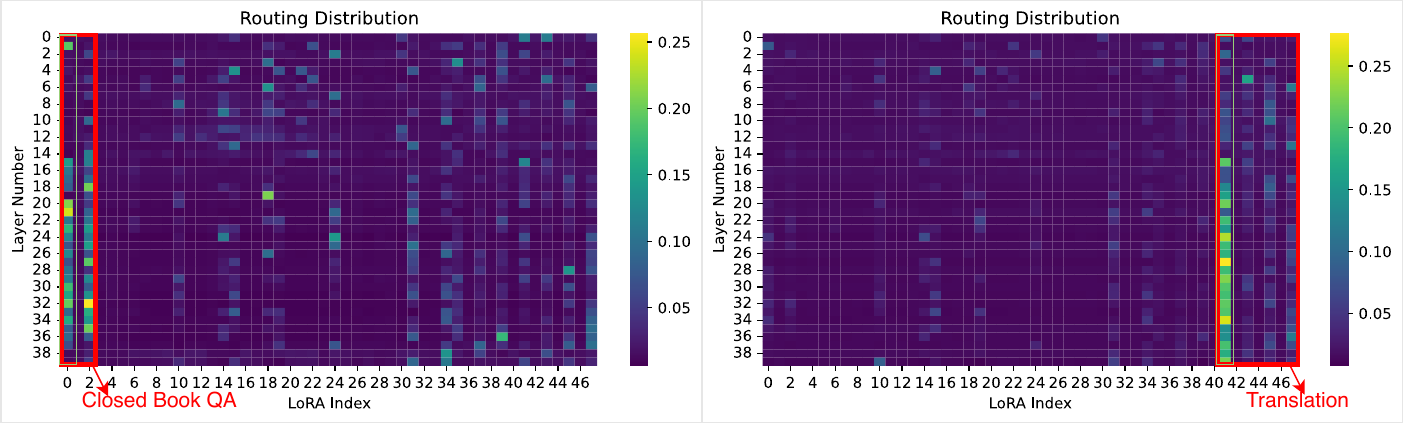}
    \caption{DAMoLE Routing Distribution for NQ (Closed-book QA) and WMT-16 De/En (Translation). Red square brackets denote LoRAs from the same task cluster and yellow square brackets highlight the ideal LoRA for the respective tasks.}
\label{fig:moe_routing}
\end{figure*}

\begin{figure}
    \centering
    \includegraphics[width= \linewidth]{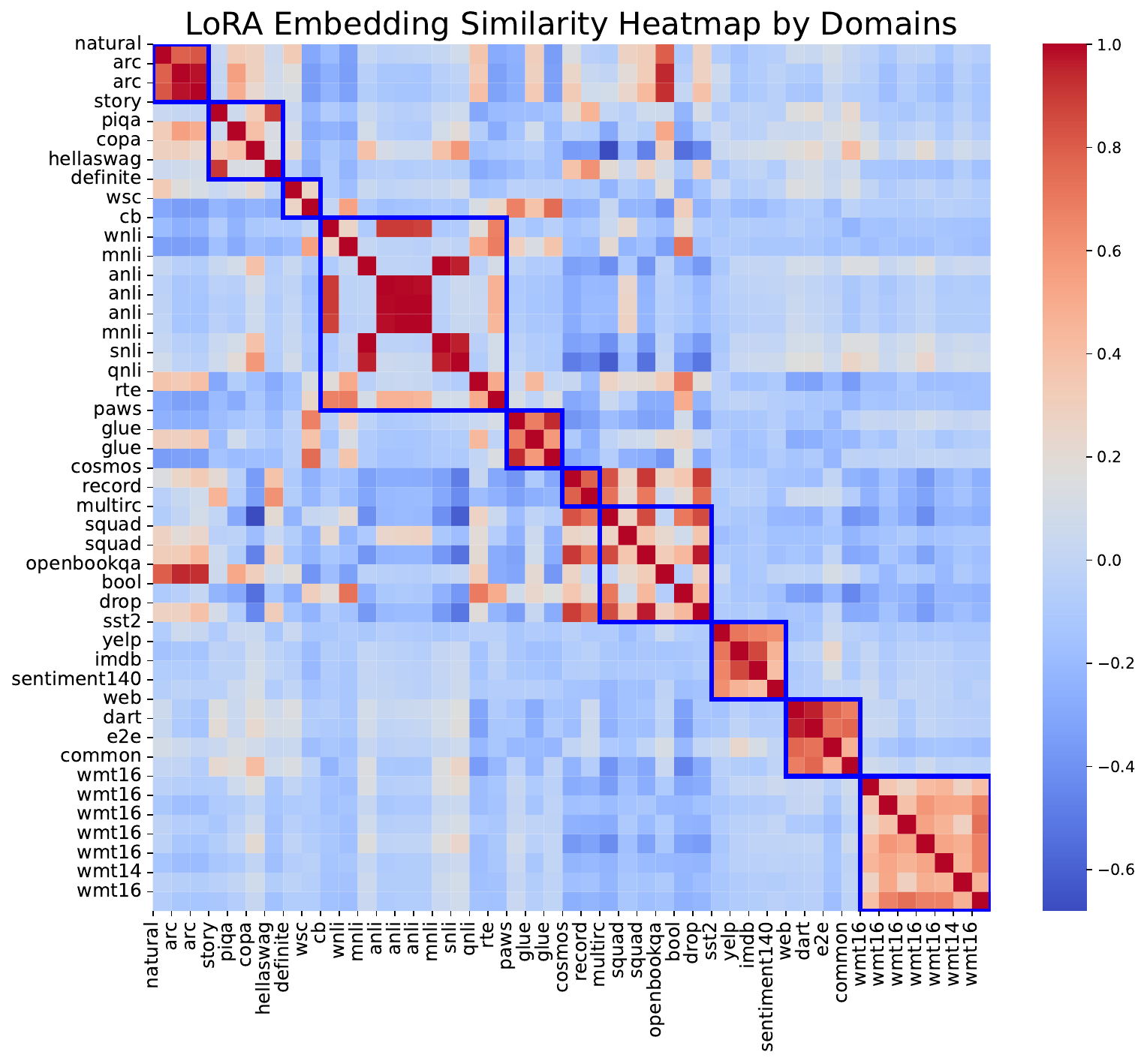}
    \caption{LoRA embedding similarity heatmap. Tasks from the same domain are grouped in square brackets.}
\label{fig:similarity_heatmap}
\end{figure}

\subsection{Main Results}
The main results of the mixed-task evaluation are shown in Tab.\ref{tab:main_sum}. We present the mean performance across each task cluster.
% and additionally evaluate the LoraRetriever's effectiveness in an out-of-domain (OOD) setting.
% In the OOD configuration, we mask the corresponding LoRA for each sample, thereby inhibiting LoraRetriever from retrieving the ideal LoRA for each sample. In this way, we can assess the cross-task generalization capability of LoraRetriever. 
From the results, we have the following observations: 
\begin{itemize}
    \item The proposed framework, DAMoLE, which performs input-aware LoRA retrieval and on-the-fly MoLE, markedly surpasses other baselines focusing on specific downstream tasks. 
    \item For the vanilla LoRA composition methods, the performance of Mixture and Selection is similar in IID scenarios, while Fusion's performance is weaker compared to the other two methods. The reasons are as follows: (i) In the IID setting, LoraRetriever can achieve strong top-1 selection, leading to similar results between Selection and the Mixture; (ii) As different tasks are inherently heterogeneous, it is inferior to directly average top-$k$ LoRA parameters in the Fusion. In the OOD setting, the Mixture exceeds the performance of the Selection, and the performance of Fusion is similar to that of the Selection. The reasons can be as follows: (i) The selection cannot retrieve the associated LoRA for the input sample in the OOD setting, leading to a significant performance drop. (ii) The Mixture can fully leverage the capabilities of similar tasks to address OOD tasks, alleviating the performance drop. 
    \item Compared to the vanilla LoRA composition strategies, DAMoLE demonstrates superior performance. (i) In the IID setting, DAMoLE outperforms the Selection method because it can route to the most suitable LoRA for each task. Additionally, DAMoLE retrieves more LoRA parameters in the initial stage, achieving better recall than the Selection method, which only selects the top-1 LoRA during the retrieval stage. (ii) In the OOD setting, DAMoLE also achieves superior performance, particularly when compared with the Mixture method. This is because DAMoLE learns to coordinate different LoRAs and differentiates each LoRA during inference. In contrast, the Mixture method simply assigns uniform weights to each LoRA, which results in suboptimal performance.
    \item The performance of the MoE and SMEAR methods is weaker than that of DAMoLE. The limitation stems from the restricted capacity of these methods for adaptation and generalization to dynamically changing environments populated with diverse LoRAs, thereby diminishing their efficacy in mixed-task scenarios.
    \item In mixed-task scenarios, although AdapterSoup uniformly searches for appropriate LoRAs for downstream tasks, the retrieved LoRAs fall short in personalization for each request, hindering their effectiveness for each specific task. 
    \item LoRAHub proves to be entirely ineffective in the mix-task scenario. First, the fusion of LoRAHub depends on randomly selected LoRAs, which may not be relevant. Second, the presence of heterogeneous tasks introduces conflicting parameter optimization directions, resulting in the total breakdown of parameter fusion.
\end{itemize}
% The full results are shown in Tab.\ref{tab:full_results}\&\ref{tab:full_results_13b}.
% We have discussed the issues with model parameter fusion in detail in the Appendix.\ref{sec:hete_fusion}.

\subsection{Analysis}
% todo: 
%  MoE routing distribution showcase; done
% IID+OOD performance; done
% Dropout strategy done
% Performance with Different LoRAs; done
% Throughput done

\paragraph{Performance of Retriever.} 
We compare LoraRetriever with some popular off-the-shelf sentence embedding models in Huggingface and adopt the model nomenclature following \cite{wolf2020transformers}. 
To analyze the effect of the percentage of tasks for training LoraRetriever, we trained three variants of LoraRetriever with different percentages.
Tab.\ref{tab:retrival_acc} shows the performance of different retrieval models for retrieving relevant LoRAs. It is shown that guiding sentence embedding models with specific prompts leads to a performance improvement in retrieval compared to common retrieval models. After instruction fine-tuning, the retriever significantly enhanced the ability to retrieve the corresponding LoRA based on the input. Conducting instruction fine-tuning on 40\% of the tasks resulted in a 2.36\% increase in top-1 accuracy and a 9.80\% increase in top-3 accuracy. Training across all tasks achieved the largest improvement. To demonstrate the generalizability of the proposed framework when dealing with unseen LoRAs, we used a retriever trained on 40\% of the tasks in the main experiment to simulate the scenario of dynamic updates to the LoRA pool that might occur while providing services with LoraRetriever.

% Tab.\ref{table:methods_improvement} shows the performance of LoraRetriever trained with different proportions of tasks for LoRA retrieval. It is observed that for the selection and mixture methods, training under 40\% of the tasks has already seen significant improvements. The best performance is achieved when trained under all tasks, but the improvement compared to 40\% is relatively small, which to some extent reflects the good generalization ability of instruction fine-tuning of LoraRetriever.

Fig.\ref{fig:similarity_heatmap} illustrates the similarity between task embeddings for different tasks through a heatmap, where tasks from the same task cluster are grouped in square brackets. It is shown that task embeddings within the same domain are more similar, indicating that the LoraRetriever embeddings can serve as task embeddings to characterize the similarities between different tasks and can be applied for LoRA retrieval.

\begin{figure}
    \centering
    \includegraphics[width= .8\linewidth]{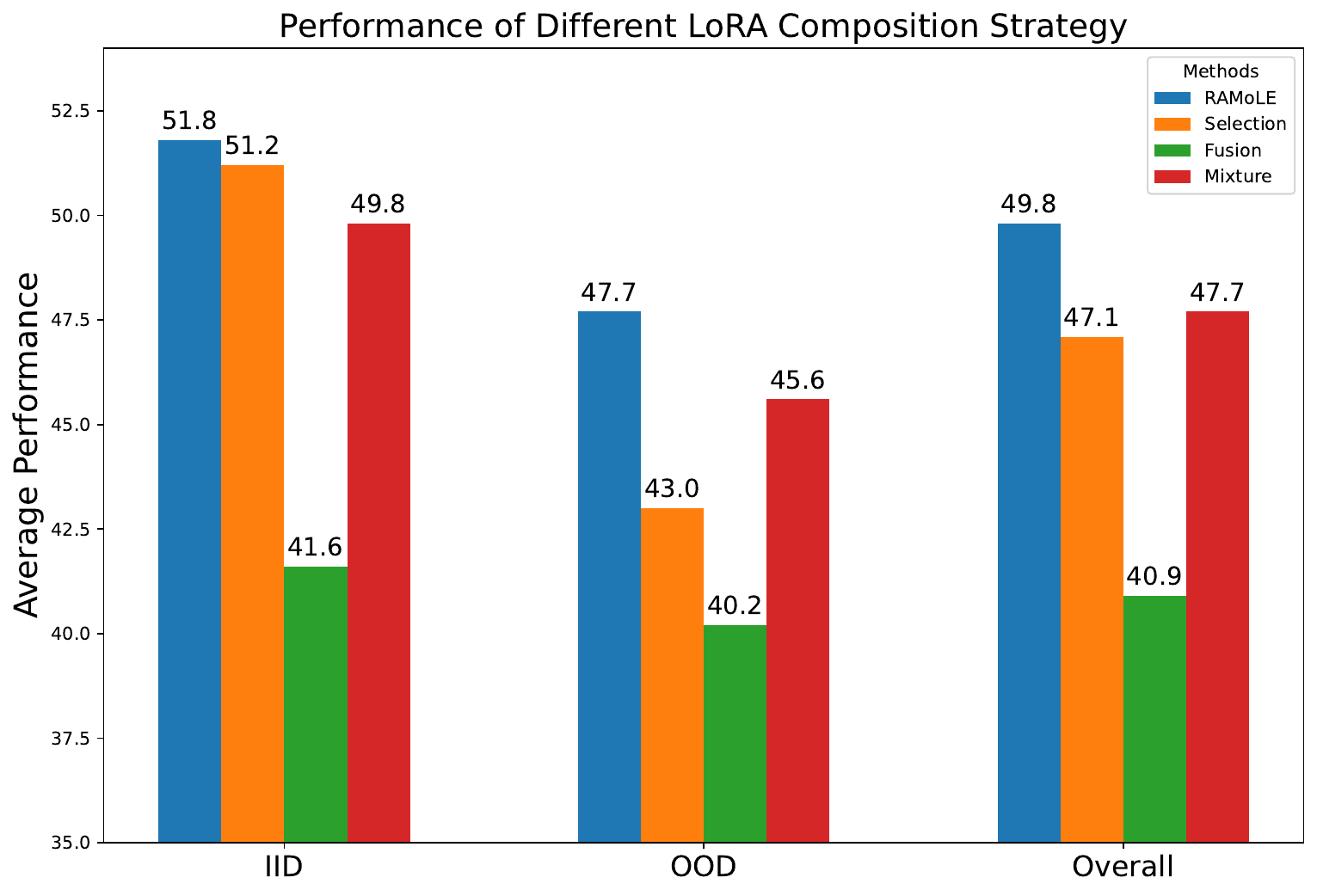}
    \caption{Comparison of Different LoRA Composition Strategies. The overall performance averages the IID and OOD results to demonstrate the comprehensive effectiveness of various LoRA composition strategies.}
\label{fig:composition_compare}
\end{figure}

\paragraph{Performance of the On-the-fly MoLE Mechanism}
Fig. \ref{fig:moe_routing} illustrates the routing distribution performance of the proposed mechanism across two tasks: Closed Book QA and Translation. The figure uses red square brackets to denote LoRAs from the same task cluster and yellow square brackets to highlight the ideal LoRA for the respective tasks. The results demonstrate that the proposed MoLE mechanism effectively assigns a higher weight to the ideal LoRA for each input at the module level. Furthermore, the importance of higher-level layers in the routing process is evident, which corroborates the findings of \cite{gao2024higher}.

In Fig.\ref{fig:composition_compare}, we compare the performance of different LoRA composition strategies across various settings. The results reveal that: (i) Among vanilla LoRA composition methods, the Selection strategy performs best in the IID setting as it effectively utilizes the ideal LoRA retrieved from LoraRetriever. The Mixture strategy excels in the OOD setting due to its ability to leverage multiple similar-task LoRAs, enhancing its generalization capability. (ii) RAMoLE achieves the best performance in both IID and OOD settings; in the IID context, it capitalizes on the optimal LoRA's strengths to compensate for the shortcomings of the Mixture strategy, while in the OOD context, it coordinates the capabilities of different LoRAs to address an unseen task effectively. Overall, RAMoLE can balance the accuracy of LoRA selection and the coordination between them, thereby achieving optimal results.

\begin{figure}
    \centering
    \includegraphics[width= .8\linewidth]{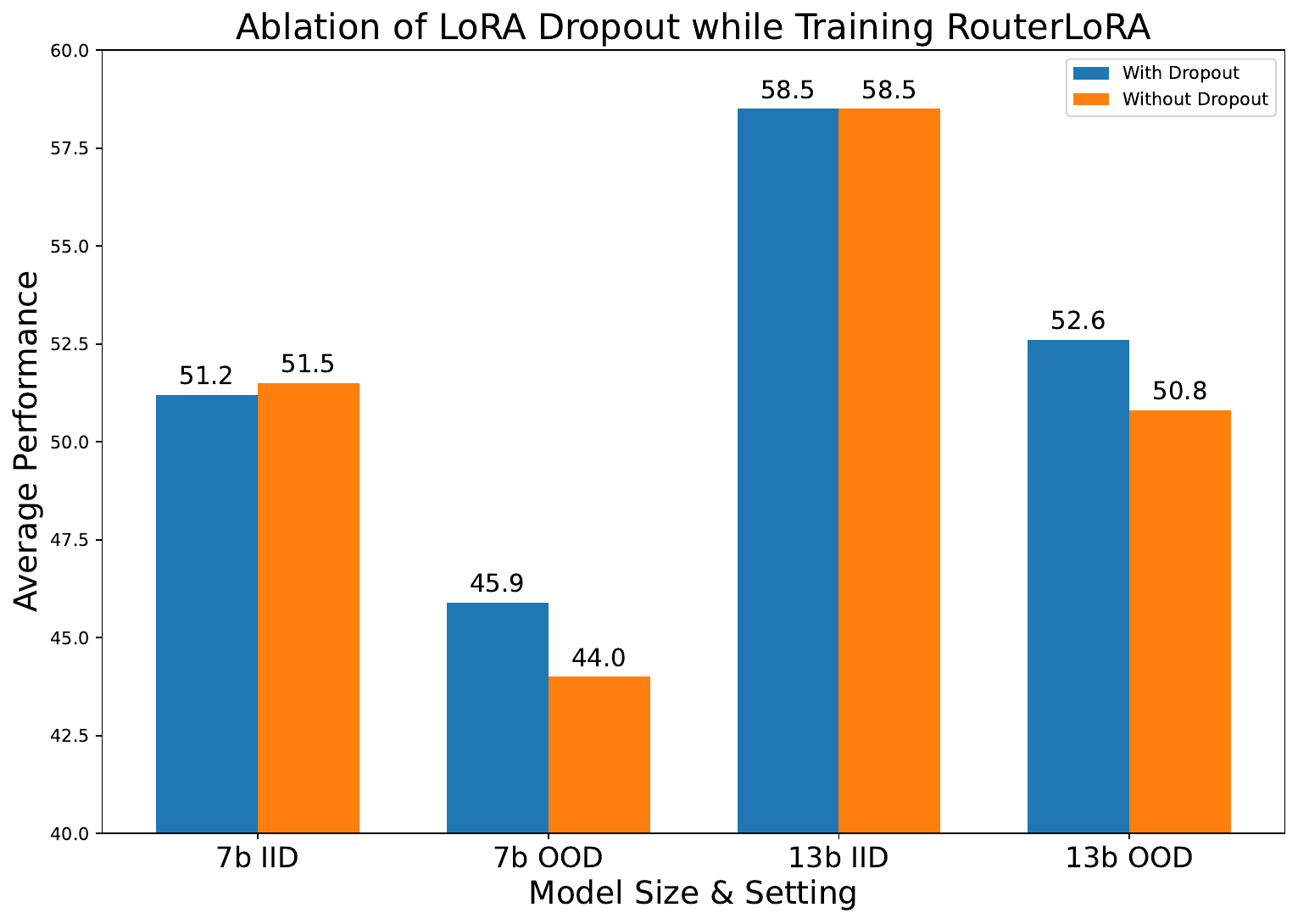}
    \caption{Ablation of LoRA Dropout Strategy during RouterLoRA training.}
\label{fig:dropout_training}
\end{figure}

We also conducted ablation experiments to demonstrate the effectiveness of the LoRA Dropout strategy during RouterLoRA training. The results are depicted in Fig.\ref{fig:dropout_training}. In the IID setting, since all training LoRAs are loaded into the LLM during training, it is easier to learn the pattern of routing to the ideal LoRA during inference. Consequently, even without LoRA Dropout, satisfactory performance is achieved in the IID setting. However, in the OOD setting, where the ideal LoRA for each sample is blocked, RouterLoRA must learn to coordinate different LoRAs to solve an unseen task. In this scenario, training without LoRA Dropout proves ineffective, and we observe a significant increase when performing LoRA Dropout in the OOD setting.

\begin{figure}
    \centering
\includegraphics[width=\linewidth]{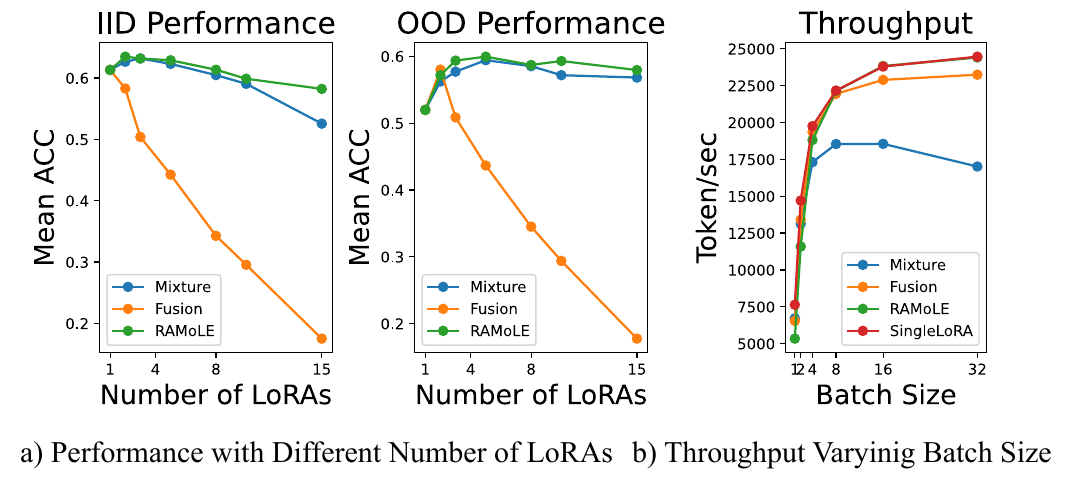}
    \caption{The left figure shows the performance of LoraRetriever varying the number of LoRAs. The right figure shows the performance of Throughput varying the batch size.}
    \label{fig:loranum}
\end{figure}

\paragraph{Impact of the number of Retrieved LoRA} Fig.\ref{fig:loranum} (a) illustrates the performance of the number of retrieved LoRAs on the mean accuracy of the NLU tasks. The results indicate that as the number of retrieved LoRAs increases, the performance of the DAMoLE and vanilla Mixture initially improves slightly but then stabilizes, while DAMoLE outperform other methods in all settings. In contrast, the Fusion shows a continuous decline in performance with an increasing number of LoRAs, which once again demonstrates that under the conditions of heterogeneous tasks, the simple averaging of parameters can compromise the original capabilities of the LoRAs. DAMoLE consistently outperforms other methods across all settings. In particular, in the OOD setting, the performance of the DAMoLE and Mixture improves significantly as the number of LoRAs increases, illustrating that in the absence of an ideal LoRA choice for a request, leveraging the capabilities of multiple LoRAs of similar tasks can effectively achieve cross-task generalization.

% \begin{figure}
%     \centering
%     % \vspace{-0.15in}
%     \subfigure[IID performance]{
% \begin{minipage}[b]{.45\linewidth}
%   \centering
%   \includegraphics[width=.85\linewidth]{fig/iid_loranum.pdf}
%     \vspace{-0.15in}
% \end{minipage}
%   }\subfigure[OOD Performance]{
% \begin{minipage}[b]{.45\linewidth}
%   \centering
%   \includegraphics[width=.85\linewidth]{fig/ood_loranum.pdf}
%     \vspace{-0.15in}
% \end{minipage}
%   }
%   \vspace{-0.15in}
%     \caption{Variation in Performance with different numbers of retrieved LoRAs. Note that when only one LoRA is retrieved, both the Mixture and Fusion methods degrade to the Selection method.}
%     \vspace{-0.15in}
% \label{fig:loranum}
% \end{figure}

% \textbf{Generalization of Retriever.}

\paragraph{Effectiveness of Batch Inference Strategy.}
To evaluate the efficiency of our proposed batch inference strategy, we compared the throughput of different batch sizes. Throughput is defined as the number of both input and output tokens per second across all requests in the mixed-task benchmark. We specifically compared the computational efficiency with that of a single LoRA. Our evaluation encompassed the entire evaluation dataset, and we limited the generation to the first produced token to mitigate discrepancies caused by varying generation lengths across different methods. These experiments were carried out on an NVIDIA A100 GPU (80GB) utilizing bfloat16 precision. As illustrated in Fig.\ref{fig:loranum} (b), our batch inference strategy markedly improves the throughput of the framework, with a slight throughput reduction compared to a single LoRA. Notably, the Fusion strategy outperforms the mixture strategy in throughput efficiency, attributed to its parameter averaging approach, which circumvents the need for parallel computation across multiple LoRAs. RAMoLE outperforms other composition methods by first gathering the corresponding LoRA for each input, allowing for a more effective inference process. As the batch size increases, the throughput of RAMoLE gradually reaches parity with that of a single LoRA.

\begin{figure}
    \centering
    \includegraphics[width=\linewidth]{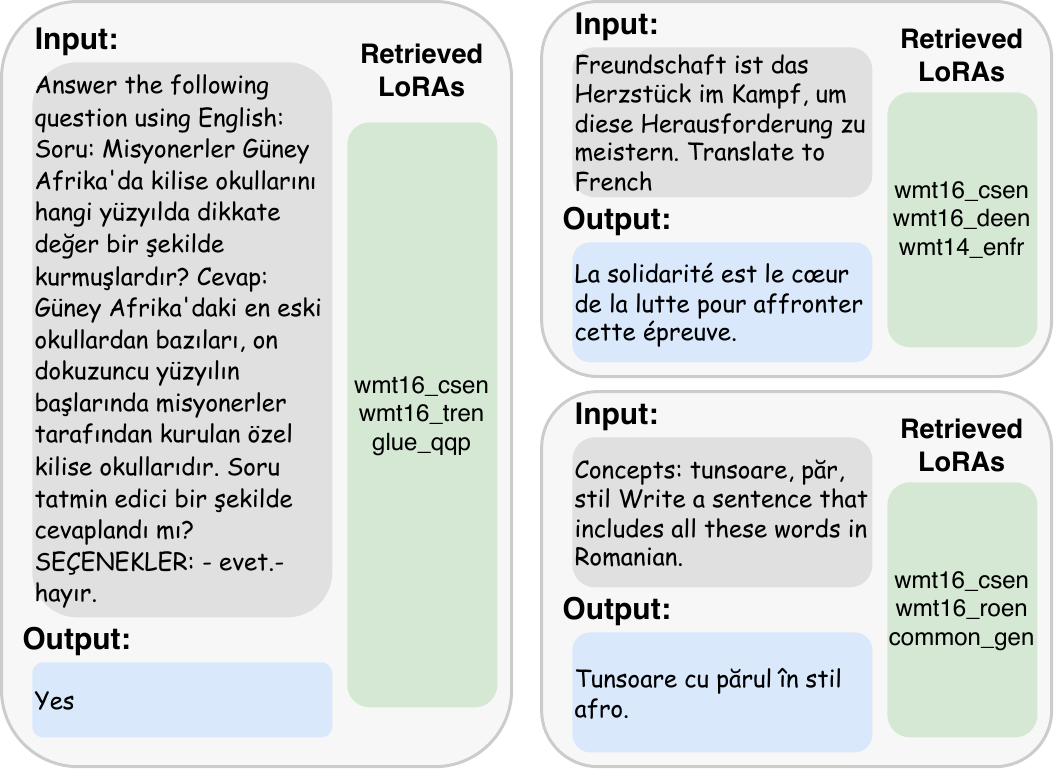}
    \caption{Showcasing How the RAMoLE Framework Employs Multiple LoRAs for Cooperative Problem Solving.}
\label{fig:lora_ability_fusion}
\end{figure}

\paragraph{Showcases.} We showcase the framework's ability to adeptly integrate multiple LoRAs for synergistic problem solving, as evidenced in Fig.\ref{fig:lora_ability_fusion}. We manually craft three problems in Fig.\ref{fig:lora_ability_fusion}, which cannot retrieve any single LoRA to solve these problems directly, necessitating the cooperation of existing LoRAs.
% challenges presented surpass the problem-solving capacity of any single LoRA, demonstrating the necessity for a cohesive blend of diverse LoRA functionalities.
Specifically, the first example requires RAMoLE to integrate NLI and translation tasks' capabilities. The retrieved LoRA wmt16-tren is utilized for comprehending Turkish, while glue-qqp is applied to NLI tasks. In the second scenario, LoRAs are integrated for translating from German to French. Although there is no direct LoRA for German-to-French translation, the combined use of wmt16-deen for German-to-English and wmt14-enfr for English-to-French enables an effective German-to-French translation. The third scenario illustrates the fusion of distinct capabilities by combining Romanian translation with text generation: leveraging the wmt16-roen LoRA for Romanian comprehension and the common-gen LoRA for generating text, RAMoLE successfully merges these diverse functionalities.
This demonstration emphasizes the framework's substantial ability to blend distinct LoRA capabilities, anticipating further exploration of capability fusion of LoRAs as a future direction.

% \begin{figure}
%     \centering
%     \includegraphics[width=.8\linewidth]{fig/throughput.pdf}
%     \caption{Tokens Processed Per Second. This figure compares the throughput of two different LoRA composition methods across varying batch sizes, evaluating the efficiency of batch inference strategies. The red line represents inference using a single LoRA at a batch size of 32, which can be viewed as the upper limit for LoRA composition methods}
%     \vspace{-0.15in}
%     \label{fig:throughput}
%     % \vspace{-0.1in}
% \end{figure}

\section{Limitations and Potential Future Directions}
While promising, there are still some drawbacks of RAMoLE. 
(1) User data privacy issues. When users upload LoRA, we need to use a small amount of training data (10-20 pieces) to represent the distribution of the LoRA model. In privacy-sensitive scenarios, representation with data may not be feasible. Aligning LoRA parameters and sample distributions in the embedding space in a manner that respects data privacy presents a worthwhile direction for future exploration.
(2) The proposed RAMoLE framework is only suitable for multi-LoRA collaboration under the same model architecture. However, in reality, the model architecture chosen by the users themselves and the PEFT method are not necessarily the same, which is worth further research on how to design the corresponding collaborative mechanism for such scenarios.

\section{Conclusion}
In this paper, we explore a new and increasingly significant computing paradigm, \textit{Uploadable Machine Learning}. The challenge in this scenario is to leverage a dynamically changing and continuously updated LoRA pool to deliver personalized services for heterogeneous downstream requests. To this end, we introduce a novel framework called RAMoLE which first identifies and retrieves suitable LoRAs based on specific prompts and then performs on-the-fly LoRA composition. The composition process incorporates a novel MoLE mechanism that uses an off-the-shelf RouterLoRA to assign weights to each LoRA. Additionally, we have developed an efficient batch inference strategy to handle batched requests. Further experiments have demonstrated the effectiveness of our proposed RAMoLE framework.
% This paper investigates a new problem of serving multiple LoRAs with a dynamically updated LoRA pool for downstream heterogeneous requests. To this end, we introduce a framework named LoraRetriver to identify and retrieve the appropriate LoRAs based on a specific input. Subsequently, we focus on the composition of these retrieved LoRAs to ensure a tailored and practical application in real-world situations. We also propose an efficient batch inference strategy to accommodate batched requests. Subsequent experiments have also demonstrated the effectiveness of our proposed LoraRetriever.

% \section*{Ethics Statement}
% In this work, we take the open-sourced Llama-2~\cite{touvron2023llama} model and Flan-v2~\cite{wei2021finetuned} dataset in the experiment. One potential ethical concern may arise when applying the proposed framework to real-world scenarios, as the data uploaded by LoRA contributors may contain user privacy. However, we believe this can be addressed by proper anonymization before uploading. Therefore, we believe our work will not pose a severe ethical concern.

\bibliography{reference}

\begin{IEEEbiography}[{\includegraphics[width=1in,height=1.25in,clip,keepaspectratio]{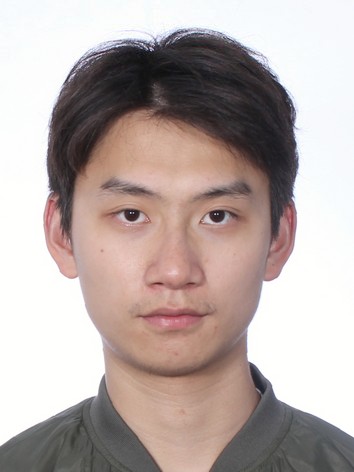}}]
{Ziyu Zhao}
received his B.S. degree in 2021 from the Department of Computer Science and Technology, Zhejiang University. He is a third-year Ph.D. candidate in the Department of Computer Science and Technology at Zhejiang University. His main research interests include Large Language Models, machine learning, and data mining.
\end{IEEEbiography}

\begin{IEEEbiography}[{\includegraphics[width=1in,height=1.25in,clip,keepaspectratio]{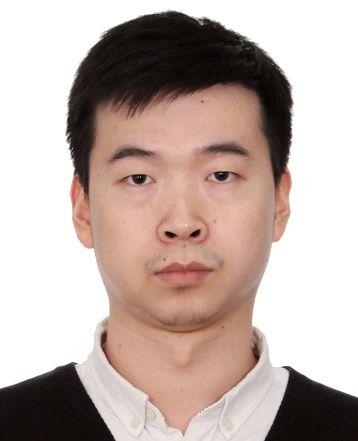}}]
{Leilei Gan}
is currently an assistant professor in the College of Software and Technology at Zhejiang University. Before that, He obtained his Ph.D. degree in computer science and technology from Zhejiang University. His research interests include natural language processing and deep learning. Currently, he studies foundation models, including pre-training, fine-tuning, alignment, and their applications on downstream tasks.
\end{IEEEbiography}

\begin{IEEEbiography}[{\includegraphics[width=1in,height=1.25in,clip,keepaspectratio]{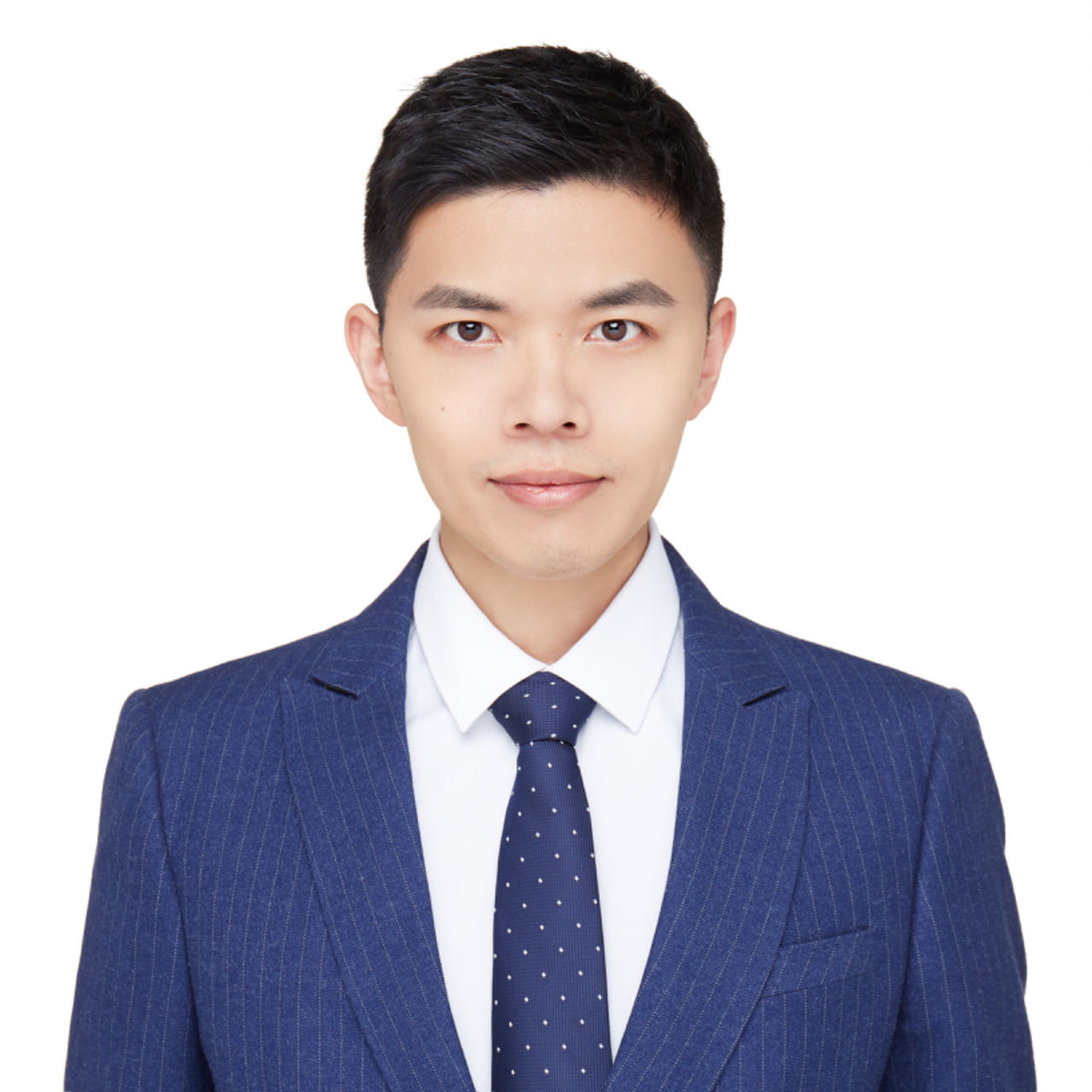}}]
{Guoyin Wang}
Dr. Guoyin Wang obtained his B.S. degree from Peking University in 2011, followed by an M.S. degree from the University of Illinois at Urbana-Champaign in 2013, and a Ph.D. from Duke University in 2020. He has authored over 50 papers in the fields of deep learning and natural language processing. His primary research interests encompass large language models, machine learning, and natural language processing. He has also served as a reviewer and area chair for prestigious academic conferences, including ACL, EMNLP, NeurIPS, and ICML.
\end{IEEEbiography}

\begin{IEEEbiography}[{\includegraphics[width=1in,height=1.25in,clip,keepaspectratio]{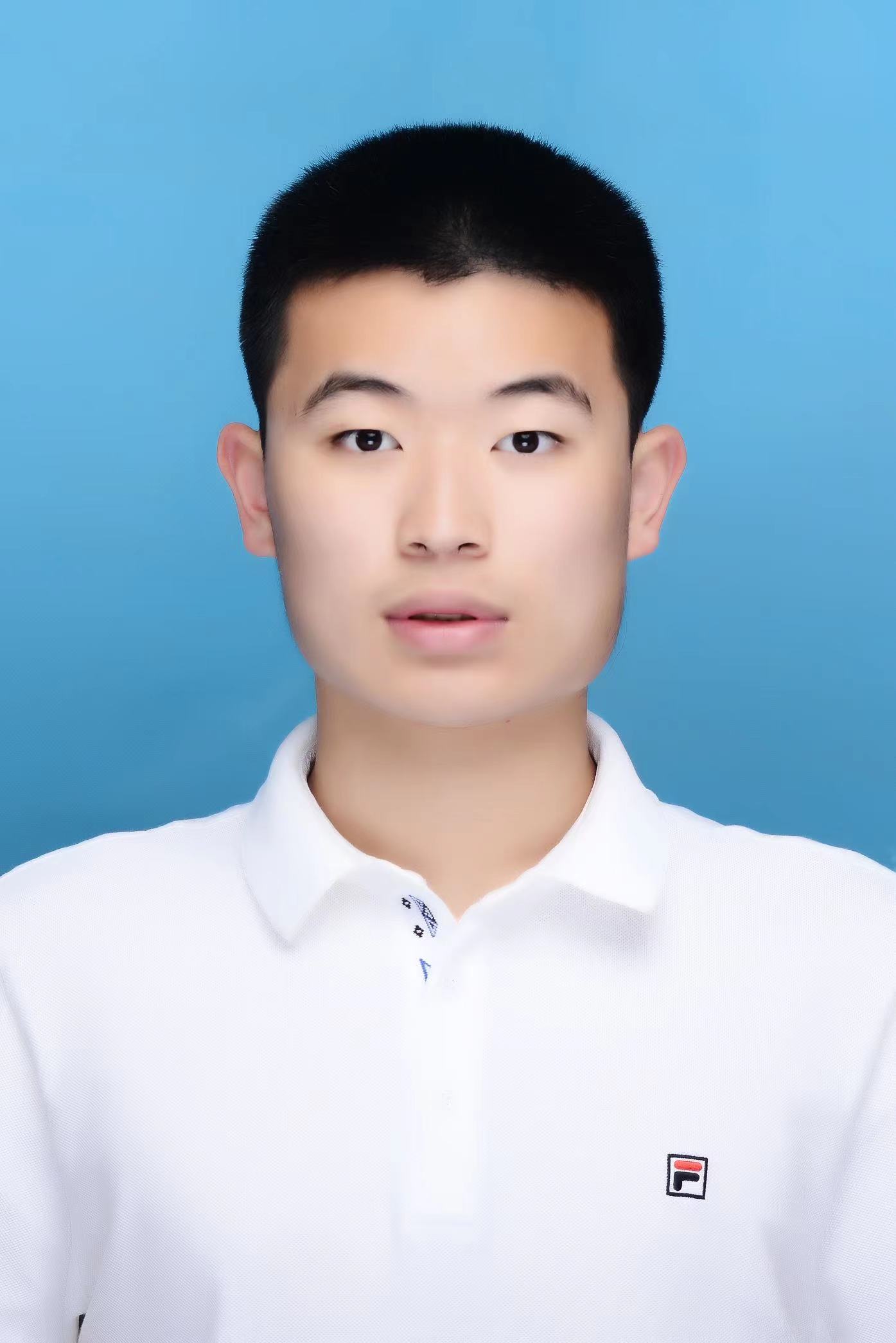}}]
{Yuwei Hu}
Yuwei Hu is a junior undergraduate student at the College of Computer Science and Technology at Zhejiang University. He is deeply engaged in the field of computer science, with a strong focus on cutting-edge technologies including large language models, machine learning, and reinforcement learning for intelligent agents.
\end{IEEEbiography}

\begin{IEEEbiography}[{\includegraphics[width=1in,height=1.25in,clip,keepaspectratio]{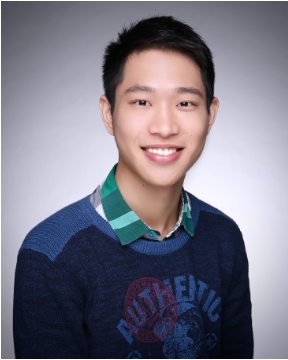}}]
{Tao Shen}
Tao Shen received his BS degree in control science and technology from China University of Petroleum, Qingdao, China, in 2015. He is currently pursuing his Ph.D. in computer science and technology at Zhejiang University. His research interests focus on federated learning.
\end{IEEEbiography}

\begin{IEEEbiography}[{\includegraphics[width=1in,height=1.25in,clip,keepaspectratio]{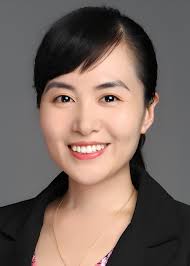}}]
{Hongxia Yang}
Dr. Hongxia Yang, PhD from Duke University, led the team to develop AI open sourced platforms and systems such as AliGraph, M6, Luoxi. Dr. Yang has published over 100 top conference and journal papers, and held more than 50 patents. She has been awarded the highest prize of the 2019 World Artificial Intelligence Conference, Super AI Leader (SAIL Award), the second prize of the 2020 National Science and Technology Progress Award, the first prize of Science and Technology Progress of the Chinese Institute of Electronics in 2021, the Forbes China Top 50 Women in Science and Technology and Ministry of Education Science and Technology Progress Award First Class in 2022 and AI 2000 Most Influential Scholar Award in 2023-2024. She used to work as the Senior Staff Data Scientist and Director in Alibaba Group, Principal Data Scientist at Yahoo! Inc and Research Staff Member at IBM T.J. Watson Research Center, joint adjunct professor at Zhejiang University Shanghai Advanced Research Institute respectively, Bytedance US Head.\end{IEEEbiography}

\begin{IEEEbiography}[{\includegraphics[width=1in,height=1.25in,clip,keepaspectratio]{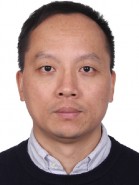}}]
{Fei Wu} received his Ph.D. degree from the College of Computer Science at Zhejiang University. Now he is the professor and dean of the College of Computer Science, Zhejiang University. His main research interests include multimedia information analysis and retrieval, and digital library.
\end{IEEEbiography}

\begin{IEEEbiography}[{\includegraphics[width=1in,height=1.25in,clip,keepaspectratio]{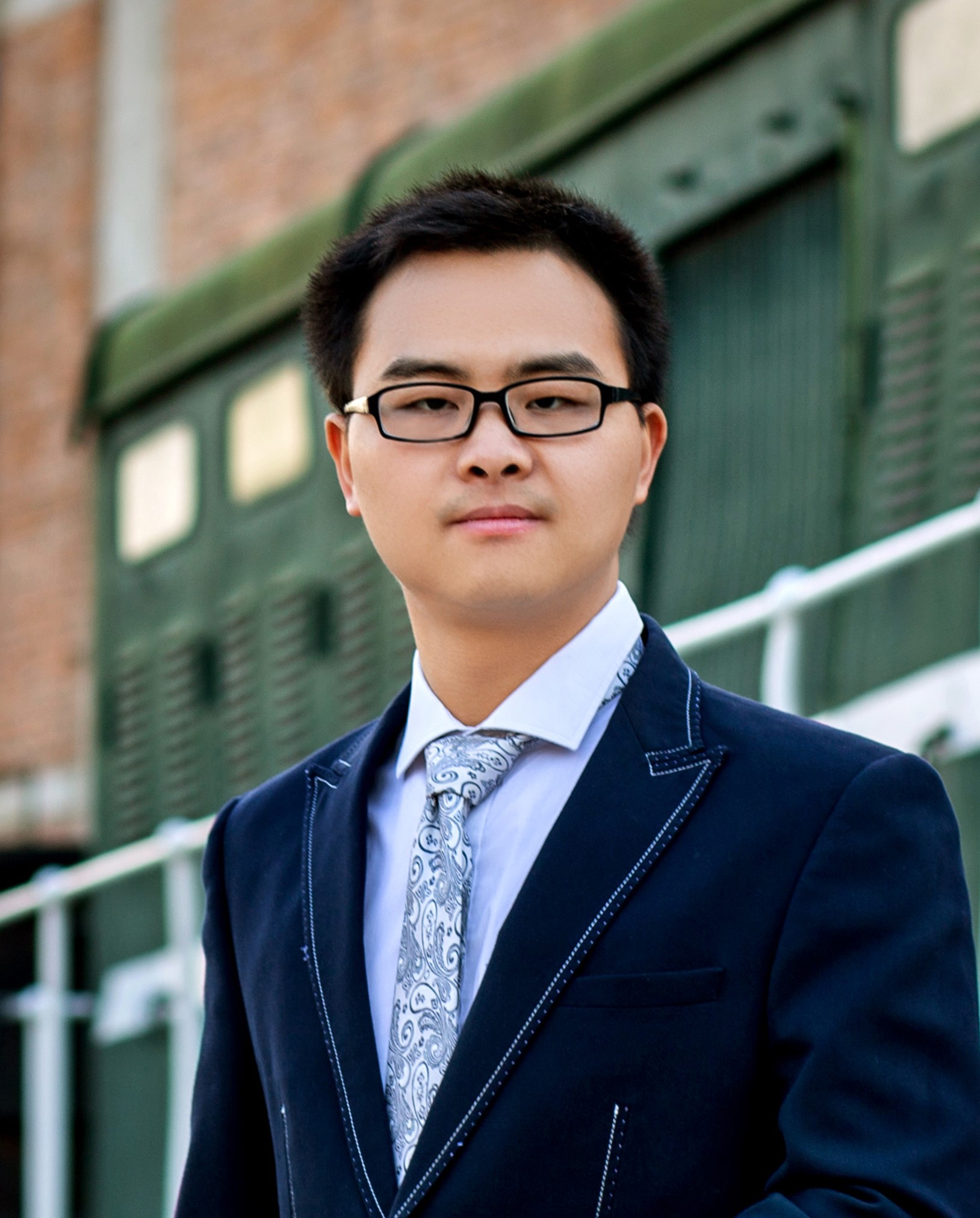}}]
{Kun Kuang} received his Ph.D. degree from Tsinghua University in 2019. He is now an Associate Professor in the College of Computer Science and Technology, Zhejiang University. He was a visiting scholar with Prof. Susan Athey's Group at Stanford University. His main research interests include Causal Inference, Artificial Intelligence, and Causally Regularized Machine Learning.  He has published over 60 papers in major international journals and conferences, including SIGKDD, ICML, NeurIPS, ACM MM, AAAI, TKDE, TKDD, Engineering, and ICDM, etc.
\end{IEEEbiography}

\newpage
\clearpage
\setcounter{page}{1}
\color{black}
\appendices
\section{Full Results}
The full results on the evaluation dataset are shown in Tab.\ref{tab:full_results} \& Tab.\ref{tab:full_results_13b}.
\begin{table*}
\caption{Mixed Tasks evaluation on both NLU \& NLG tasks. ``OOD" indicates that during retrieval, we masked the corresponding task's LoRA for testing generalization when facing unknown tasks.}
\centering
\resizebox{.8\linewidth}{!}{  
\begin{tabular}{l c c c c c c c c c c c c c c c} 
\toprule
\multirow{2}{*}{Task} &
\multirow{2}{*}{\begin{tabular}[c]{@{}c@{}}Perfect\\Selection\end{tabular}} &
\multicolumn{2}{c}{RAMoLE} & 
\multicolumn{2}{c}{Selection} & 
\multicolumn{2}{c}{Fusion} & 
\multicolumn{2}{c}{Mixture} & 
\multirow{2}{*}{\begin{tabular}[c]{@{}c@{}}MoE\\Top1\end{tabular}} &
\multirow{2}{*}{\begin{tabular}[c]{@{}c@{}}MoE\\Top3\end{tabular}} &
\multirow{2}{*}{\begin{tabular}[c]{@{}c@{}}MoE\\Soft\end{tabular}} &
\multirow{2}{*}{\begin{tabular}[c]{@{}c@{}}SME-\\AR\end{tabular}} &
\multirow{2}{*}{\begin{tabular}[c]{@{}c@{}}Adapter\\Soup\end{tabular}} &
\multirow{2}{*}{\begin{tabular}[c]{@{}c@{}}LoRA\\Hub\end{tabular}}\\
\cline{3-10}
& & IID & OOD & IID & OOD & IID & OOD & IID & OOD & & & & \\
\midrule
\multicolumn{12}{l}{\textbf{Struct to Text}} \\
WebNLG \textsuperscript{Rouge-1} & 68.2 & 51.9 & 46.8 & 64.2 & 50.5 & 45.8 & 41.3 & 54.6 & 50.1 & 41.9 & 43.2 & 45.0 & 46.7 & 3.1 & 29.6 \\
WebNLG \textsuperscript{Rouge-2} & 47.6 & 25.1 & 23.5 & 41.6 & 26.3 & 23.2 & 21.1 & 30.1 & 25.9 & 20.1 & 22.2 & 22.9 & 25.2 & 0.9 & 14.7 \\
WebNLG \textsuperscript{Rouge-l} & 56.6 & 40.7 & 36.8 & 51.6 & 39.7 & 36.3 & 32.3 & 44.6 & 40.9 & 32.9 & 33.0 & 35.1 & 36.0 & 2.7 & 23.9 \\
DART \textsuperscript{Rouge-1} & 66.6 & 65.8 & 50.6 & 63.1 & 52.0 & 51.1 & 48.8 & 58.2 & 54.8 & 51.7 & 52.4 & 52.5 & 56.0 & 1.9 & 36.7 \\
DART \textsuperscript{Rouge-2} & 45.3 & 43.1 & 29.7 & 42.1 & 30.5 & 28.5 & 26.9 & 32.7 & 31.4 & 28.5 & 28.6 & 28.1 & 30.3 & 0.5 & 18.2 \\
DART \textsuperscript{Rouge-l} & 55.1 & 54.3 & 39.9 & 52.7 & 42.6 & 39.9 & 37.6 & 46.7 & 43.8 & 41.8 & 42.1 & 41.3 & 45.1 & 1.7 & 29.0 \\
E2ENLG \textsuperscript{Rouge-1} & 59.2 & 58.5 & 56.0 & 58.8 & 53.2 & 54.7 & 50.4 & 59.1 & 52.3 & 46.8 & 47.7 & 49.3 & 47.3 & 4.0 & 43.7 \\
E2ENLG \textsuperscript{Rouge-2} & 34.3 & 32.9 & 30.6 & 33.6 & 29.6 & 29.8 & 27.1 & 33.2 & 27.6 & 23.0 & 23.6 & 25.0 & 23.4 & 2.3 & 21.9 \\
E2ENLG \textsuperscript{Rouge-l} & 44.8 & 44.1 & 41.3 & 44.4 & 38.6 & 39.8 & 37.7 & 43.9 & 37.1 & 34.5 & 35.2 & 37.0 & 34.7 & 3.9 & 32.0 \\
CommonGen \textsuperscript{Rouge-1} & 42.4 & 42.9 & 27.5 & 41.1 & 25.1 & 26.4 & 23.3 & 32.8 & 24.0 & 24.7 & 24.0 & 25.9 & 23.3 & 5.2 & 17.6 \\
CommonGen \textsuperscript{Rouge-2} & 17.2 & 18.7 & 8.6 & 16.9 & 6.3 & 8.8 & 5.6 & 9.3 & 6.6 & 5.5 & 5.0 & 6.9 & 6.4 & 0.0 & 5.6 \\
CommonGen \textsuperscript{Rouge-l} & 37.8 & 39.6 & 24.1 & 37.0 & 20.4 & 21.9 & 19.1 & 28.7 & 20.4 & 21.1 & 21.0 & 21.9 & 19.8 & 4.9 & 14.8 \\
\midrule
\multicolumn{12}{l}{\textbf{Translation}} \\
WMT'16-tren & 3.2 & 2.7 & 2.0 & 3.1 & 2.6 & 3.5 & 3.3 & 3.7 & 2.6 & 3.5 & 3.2 & 3.4 & 3.2 & 0.0 & 2.0 \\
WMT'16-deen  & 18.9 & 18.4 & 18.4 & 18.7 & 20.3 & 17.9 & 18.8 & 18.8 & 18.7 & 11.6 & 14.0 & 14.7 & 16.6 & 1.1 & 11.4 \\
WMT'16-ruen & 10.8 & 9.9 & 10.3 & 10.4 & 9.8 & 9.2 & 9.3 & 11.0 & 10.8 & 6.2 & 7.8 & 8.3 & 7.3 & 0.0 & 4.8 \\
WMT'16-fien & 6.5 & 7.6 & 7.3 & 6.5 & 7.0 & 7.2 & 7.1 & 7.3 & 7.8 & 6.2 & 6.2 & 6.1 & 6.5 & 0.7 & 4.3 \\
WMT'16-roen & 13.9 & 15.9 & 12.9 & 14.0 & 12.3 & 12.8 & 13.3 & 13.1 & 12.2 & 9.8 & 10.7 & 10.1 & 10.3 & 0.3 & 8.0 \\
WMT'14-enfr  & 16.5 & 16.2 & 17.1 & 16.1 & 16.9 & 17.7 & 18.0 & 17.8 & 18.0 & 15.9 & 17.3 & 17.1 & 16.4 & 3.5 & 15.2 \\
WMT'16-csen & 10.7 & 7.0 & 5.2 & 9.4 & 7.0 & 6.1 & 6.2 & 8.3 & 5.8 & 4.7 & 6.3 & 6.3 & 6.3 & 0.8 & 6.1 \\
Paracrawl-enes & 24.3 & 23.2 & 21.3 & 24.2 & 20.3 & 22.9 & 22.3 & 22.8 & 22.1 & 18.0 & 18.8 & 19.5 & 21.6 & 4.5 & 16.4 \\
\midrule
\multicolumn{12}{l}{ \textbf{COMMONSENSE}}\\
StoryCloze & 72.0 & 78.0 & 90.0 & 62.0 & 42.0 & 72.0 & 68.0 & 84.0 & 58.0 & 74.0 & 70.0 & 70.0 & 68.0 & 62.0 & 48.0 \\
PIQA & 46.0 & 44.0 & 38.0 & 46.0 & 32.0 & 34.0 & 36.0 & 38.0 & 34.0 & 40.0 & 38.0 & 38.0 & 36.0 & 38.0 & 0.0 \\
COPA & 86.0 & 62.0 & 58.0 & 74.0 & 68.0 & 78.0 & 70.0 & 80.0 & 68.0 & 72.0 & 70.0 & 72.0 & 70.0 & 56.0 & 22.0 \\
HellaSwag & 46.0 & 36.0 & 32.0 & 40.0 & 42.0 & 20.0 & 18.0 & 44.0 & 40.0 & 32.0 & 30.0 & 26.0 & 26.0 & 28.0 & 0.0 \\
\hline
\multicolumn{12}{l}{\textbf{sentiment}}    \\
SST-2 & 98.0 & 96.0 & 96.0 & 98.0 & 96.0 & 74.0 & 78.0 & 96.0 & 94.0 & 56.0 & 68.0 & 66.0 & 66.0 & 74.0 & 0.0 \\
Yelp & 98.0 & 98.0 & 98.0 & 94.0 & 94.0 & 96.0 & 96.0 & 98.0 & 98.0 & 86.0 & 90.0 & 86.0 & 84.0 & 80.0 & 0.0 \\
IMDB & 96.0 & 98.0 & 100.0 & 96.0 & 96.0 & 92.0 & 82.0 & 96.0 & 96.0 & 76.0 & 80.0 & 80.0 & 84.0 & 80.0 & 0.0 \\
sentiment140 & 68.0 & 70.0 & 74.0 & 70.0 & 70.0 & 54.0 & 58.0 & 68.0 & 74.0 & 62.0 & 62.0 & 66.0 & 62.0 & 60.0 & 2.0 \\
\hline
\multicolumn{12}{l}{\textbf{READING Comp.}}\\
MultiRC & 68.0 & 64.0 & 60.0 & 52.0 & 38.0 & 44.0 & 44.0 & 48.0 & 44.0 & 54.0 & 52.0 & 50.0 & 48.0 & 40.0 & 6.0 \\
SQuADv2 & 62.0 & 26.0 & 8.0 & 56.0 & 12.0 & 30.0 & 20.0 & 22.0 & 16.0 & 24.0 & 24.0 & 26.0 & 22.0 & 16.0 & 0.0 \\
SQuADv1 & 68.0 & 62.0 & 62.0 & 66.0 & 68.0 & 64.0 & 64.0 & 62.0 & 68.0 & 68.0 & 70.0 & 66.0 & 66.0 & 54.0 & 4.0 \\
OBQA & 82.0 & 78.0 & 68.0 & 68.0 & 58.0 & 64.0 & 60.0 & 78.0 & 66.0 & 62.0 & 64.0 & 66.0 & 60.0 & 40.0 & 0.0 \\
BoolQ & 84.0 & 82.0 & 74.0 & 60.0 & 60.0 & 68.0 & 70.0 & 80.0 & 76.0 & 74.0 & 68.0 & 76.0 & 70.0 & 72.0 & 6.0 \\
drop & 40.0 & 18.0 & 16.0 & 8.0 & 6.0 & 14.0 & 12.0 & 18.0 & 14.0 & 10.0 & 8.0 & 8.0 & 8.0 & 22.0 & 0.0 \\
\hline
\multicolumn{12}{l}{\textbf{CLOSE-BOOK QA}} \\
NQ & 18.0 & 12.0 & 10.0 & 16.0 & 10.0 & 16.0 & 14.0 & 16.0 & 10.0 & 12.0 & 12.0 & 12.0 & 4.0 & 12.0 & 0.0 \\
ARC-e & 50.0 & 74.0 & 80.0 & 56.0 & 70.0 & 54.0 & 56.0 & 66.0 & 82.0 & 58.0 & 58.0 & 60.0 & 58.0 & 48.0 & 0.0 \\
ARC-c & 46.0 & 46.0 & 48.0 & 42.0 & 46.0 & 34.0 & 34.0 & 50.0 & 46.0 & 46.0 & 42.0 & 42.0 & 42.0 & 24.0 & 0.0 \\
TriviaQa & 66.0 & 56.0 & 56.0 & 46.0 & 46.0 & 60.0 & 46.0 & 48.0 & 56.0 & 46.0 & 42.0 & 46.0 & 24.0 & 42.0 & 4.0 \\
\hline
\multicolumn{12}{l}{\textbf{COREFERENCE}}\\
DPR & 54.0 & 64.0 & 46.0 & 50.0 & 50.0 & 56.0 & 60.0 & 68.0 & 56.0 & 64.0 & 60.0 & 62.0 & 62.0 & 46.0 & 2.0 \\
WSC & 50.0 & 50.0 & 56.0 & 50.0 & 42.0 & 38.0 & 46.0 & 58.0 & 42.0 & 58.0 & 58.0 & 52.0 & 54.0 & 40.0 & 0.0 \\
\hline
\multicolumn{12}{l}{\textbf{READ. COOMP. W/ COMMONSENSE}}  \\
CosmosQa & 68.0 & 74.0 & 58.0 & 68.0 & 34.0 & 46.0 & 32.0 & 50.0 & 46.0 & 44.0 & 46.0 & 44.0 & 38.0 & 14.0 & 6.0 \\
record & 70.0 & 54.0 & 30.0 & 70.0 & 26.0 & 24.0 & 6.0 & 42.0 & 34.0 & 18.0 & 12.0 & 14.0 & 8.0 & 14.0 & 0.0 \\
\hline
\multicolumn{12}{l}{\textbf{PARAPHRASE}}    \\
Paws Wiki & 90.0 & 54.0 & 44.0 & 64.0 & 40.0 & 44.0 & 42.0 & 56.0 & 46.0 & 56.0 & 50.0 & 48.0 & 54.0 & 60.0 & 2.0 \\
QQP & 74.0 & 88.0 & 64.0 & 74.0 & 68.0 & 66.0 & 60.0 & 80.0 & 58.0 & 50.0 & 40.0 & 36.0 & 28.0 & 54.0 & 0.0 \\
MRPC & 60.0 & 62.0 & 60.0 & 58.0 & 58.0 & 60.0 & 62.0 & 60.0 & 58.0 & 42.0 & 44.0 & 40.0 & 42.0 & 60.0 & 2.0 \\
STSB & 38.0 & 34.0 & 8.0 & 36.0 & 16.0 & 12.0 & 12.0 & 30.0 & 20.0 & 20.0 & 20.0 & 20.0 & 14.0 & 12.0 & 0.0 \\
\hline
\multicolumn{12}{l}{\textbf{NLI}}\\
CB & 88.9 & 82.2 & 66.7 & 80.0 & 62.2 & 77.8 & 57.8 & 86.7 & 66.7 & 68.9 & 64.4 & 68.9 & 62.2 & 55.6 & 13.3 \\
WNLI & 70.0 & 52.0 & 42.0 & 68.0 & 46.0 & 44.0 & 50.0 & 60.0 & 54.0 & 56.0 & 56.0 & 42.0 & 44.0 & 52.0 & 0.0 \\
ANLI-r1 & 50.0 & 52.0 & 48.0 & 50.0 & 50.0 & 40.0 & 42.0 & 40.0 & 42.0 & 40.0 & 40.0 & 36.0 & 38.0 & 38.0 & 24.0 \\
ANLI-r2 & 46.0 & 48.0 & 46.0 & 46.0 & 46.0 & 32.0 & 36.0 & 46.0 & 46.0 & 40.0 & 36.0 & 38.0 & 32.0 & 46.0 & 20.0 \\
ANLI-r3 & 46.0 & 50.0 & 38.0 & 42.0 & 38.0 & 38.0 & 40.0 & 44.0 & 50.0 & 28.0 & 32.0 & 34.0 & 38.0 & 40.0 & 24.0 \\
MNLI-m & 88.0 & 76.0 & 84.0 & 84.0 & 88.0 & 62.0 & 66.0 & 80.0 & 88.0 & 48.0 & 54.0 & 50.0 & 56.0 & 76.0 & 0.0 \\
MNLI-mm & 92.0 & 84.0 & 90.0 & 90.0 & 94.0 & 64.0 & 82.0 & 88.0 & 90.0 & 48.0 & 48.0 & 50.0 & 60.0 & 84.0 & 2.0 \\
SNLI & 96.0 & 88.0 & 86.0 & 84.0 & 84.0 & 56.0 & 58.0 & 90.0 & 92.0 & 54.0 & 52.0 & 54.0 & 54.0 & 82.0 & 0.0 \\
QNLI & 94.0 & 66.0 & 62.0 & 94.0 & 26.0 & 46.0 & 48.0 & 74.0 & 38.0 & 56.0 & 56.0 & 54.0 & 60.0 & 70.0 & 0.0 \\
RTE & 52.0 & 68.0 & 64.0 & 62.0 & 72.0 & 54.0 & 58.0 & 70.0 & 76.0 & 64.0 & 58.0 & 56.0 & 64.0 & 80.0 & 22.0 \\
\bottomrule
\end{tabular}
}
\label{tab:full_results}
\end{table*}

\begin{table*}
\caption{Mixed Tasks evaluation on both NLU \& NLG tasks. ``OOD" indicates that during retrieval, we masked the corresponding task's LoRA for testing generalization when facing unknown tasks.}
\centering
\resizebox{.8\linewidth}{!}{  
\begin{tabular}{l c c c c c c c c c c c c c c c} 
\toprule
\multirow{2}{*}{Task} &
\multirow{2}{*}{\begin{tabular}[c]{@{}c@{}}Perfect\\Selection\end{tabular}} &
\multicolumn{2}{c}{RAMoLE} & 
\multicolumn{2}{c}{Selection} & 
\multicolumn{2}{c}{Fusion} & 
\multicolumn{2}{c}{Mixture} & 
\multirow{2}{*}{\begin{tabular}[c]{@{}c@{}}MoE\\Top1\end{tabular}} &
\multirow{2}{*}{\begin{tabular}[c]{@{}c@{}}MoE\\Top3\end{tabular}} &
\multirow{2}{*}{\begin{tabular}[c]{@{}c@{}}MoE\\Soft\end{tabular}} &
\multirow{2}{*}{\begin{tabular}[c]{@{}c@{}}SME-\\AR\end{tabular}} &
\multirow{2}{*}{\begin{tabular}[c]{@{}c@{}}Adapter\\Soup\end{tabular}} &
\multirow{2}{*}{\begin{tabular}[c]{@{}c@{}}LoRA\\Hub\end{tabular}}\\
\cline{3-10}
& & IID & OOD & IID & OOD & IID & OOD & IID & OOD & & & & \\
\midrule
\multicolumn{12}{l}{\textbf{Struct to Text}} \\
WebNLG \textsuperscript{Rouge-1} & 68.2 & 51.9 & 46.8 & 64.2 & 50.5 & 45.8 & 41.3 & 54.6 & 50.1 & 41.9 & 43.2 & 45.0 & 46.7 & 3.1 & 29.6 \\
WebNLG \textsuperscript{Rouge-2} & 47.6 & 25.1 & 23.5 & 41.6 & 26.3 & 23.2 & 21.1 & 30.1 & 25.9 & 20.1 & 22.2 & 22.9 & 25.2 & 0.9 & 14.7 \\
WebNLG \textsuperscript{Rouge-l} & 56.6 & 40.7 & 36.8 & 51.6 & 39.7 & 36.3 & 32.3 & 44.6 & 40.9 & 32.9 & 33.0 & 35.1 & 36.0 & 2.7 & 23.9 \\
DART \textsuperscript{Rouge-1} & 66.6 & 65.8 & 50.6 & 63.1 & 52.0 & 51.1 & 48.8 & 58.2 & 54.8 & 51.7 & 52.4 & 52.5 & 56.0 & 1.9 & 36.7 \\
DART \textsuperscript{Rouge-2}  & 45.3 & 43.1 & 29.7 & 42.1 & 30.5 & 28.5 & 26.9 & 32.7 & 31.4 & 28.5 & 28.6 & 28.1 & 30.3 & 0.5 & 18.2 \\
DART \textsuperscript{Rouge-l}  & 55.1 & 54.3 & 39.9 & 52.7 & 42.6 & 39.9 & 37.6 & 46.7 & 43.8 & 41.8 & 42.1 & 41.3 & 45.1 & 1.7 & 29.0 \\
E2ENLG \textsuperscript{Rouge-1} & 59.2 & 58.5 & 56.0 & 58.8 & 53.2 & 54.7 & 50.4 & 59.1 & 52.3 & 46.8 & 47.7 & 49.3 & 47.3 & 4.0 & 43.7 \\
E2ENLG \textsuperscript{Rouge-2} & 34.3 & 32.9 & 30.6 & 33.6 & 29.6 & 29.8 & 27.1 & 33.2 & 27.6 & 23.0 & 23.6 & 25.0 & 23.4 & 2.3 & 21.9 \\
E2ENLG \textsuperscript{Rouge-l} & 44.8 & 44.1 & 41.3 & 44.4 & 38.6 & 39.8 & 37.7 & 43.9 & 37.1 & 34.5 & 35.2 & 37.0 & 34.7 & 3.9 & 32.0 \\
CommonGen \textsuperscript{Rouge-1} & 42.4 & 42.9 & 27.5 & 41.1 & 25.1 & 26.4 & 23.3 & 32.8 & 24.0 & 24.7 & 24.0 & 25.9 & 23.3 & 5.2 & 17.6 \\
CommonGen \textsuperscript{Rouge-2} & 17.2 & 18.7 & 8.6 & 16.9 & 6.3 & 8.8 & 5.6 & 9.3 & 6.6 & 5.5 & 5.0 & 6.9 & 6.4 & 0.0 & 5.6 \\
CommonGen \textsuperscript{Rouge-l} & 37.8 & 39.6 & 24.1 & 37.0 & 20.4 & 21.9 & 19.1 & 28.7 & 20.4 & 21.1 & 21.0 & 21.9 & 19.8 & 4.9 & 14.8 \\
\midrule
\multicolumn{12}{l}{\textbf{Translation}} \\
WMT'16-tren & 3.2 & 2.7 & 2.0 & 3.1 & 2.6 & 3.5 & 3.3 & 3.7 & 2.6 & 3.5 & 3.2 & 3.4 & 3.2 & 0.0 & 2.0 \\
WMT'16-deen & 18.9 & 18.4 & 18.4 & 18.7 & 20.3 & 17.9 & 18.8 & 18.8 & 18.7 & 11.6 & 14.0 & 14.7 & 16.6 & 1.1 & 11.4 \\
WMT'16-ruen & 10.8 & 9.9 & 10.3 & 10.4 & 9.8 & 9.2 & 9.3 & 11.0 & 10.8 & 6.2 & 7.8 & 8.3 & 7.3 & 0.0 & 4.8 \\
WMT'16-fien  & 6.5 & 7.6 & 7.3 & 6.5 & 7.0 & 7.2 & 7.1 & 7.3 & 7.8 & 6.2 & 6.2 & 6.1 & 6.5 & 0.7 & 4.3 \\
WMT'16-roen & 13.9 & 15.9 & 12.9 & 14.0 & 12.3 & 12.8 & 13.3 & 13.1 & 12.2 & 9.8 & 10.7 & 10.1 & 10.3 & 0.3 & 8.0 \\
WMT'14-enfr & 16.5 & 16.2 & 17.1 & 16.1 & 16.9 & 17.7 & 18.0 & 17.8 & 18.0 & 15.9 & 17.3 & 17.1 & 16.4 & 3.5 & 15.2 \\
WMT'16-csen  & 10.7 & 7.0 & 5.2 & 9.4 & 7.0 & 6.1 & 6.2 & 8.3 & 5.8 & 4.7 & 6.3 & 6.3 & 6.3 & 0.8 & 6.1 \\
Paracrawl-enes & 24.3 & 23.2 & 21.3 & 24.2 & 20.3 & 22.9 & 22.3 & 22.8 & 22.1 & 18.0 & 18.8 & 19.5 & 21.6 & 4.5 & 16.4 \\
\midrule
\multicolumn{12}{l}{ \textbf{COMMONSENSE}}\\
StoryCloze & 72.0 & 78.0 & 90.0 & 62.0 & 42.0 & 72.0 & 68.0 & 84.0 & 58.0 & 74.0 & 70.0 & 70.0 & 68.0 & 62.0 & 48.0 \\
PIQA & 46.0 & 44.0 & 38.0 & 46.0 & 32.0 & 34.0 & 36.0 & 38.0 & 34.0 & 40.0 & 38.0 & 38.0 & 36.0 & 38.0 & 0.0 \\
COPA & 86.0 & 62.0 & 58.0 & 74.0 & 68.0 & 78.0 & 70.0 & 80.0 & 68.0 & 72.0 & 70.0 & 72.0 & 70.0 & 56.0 & 22.0 \\
HellaSwag & 46.0 & 36.0 & 32.0 & 40.0 & 42.0 & 20.0 & 18.0 & 44.0 & 40.0 & 32.0 & 30.0 & 26.0 & 26.0 & 28.0 & 0.0 \\
\hline
\multicolumn{12}{l}{\textbf{sentiment}}    \\
SST-2 & 98.0 & 96.0 & 96.0 & 98.0 & 96.0 & 74.0 & 78.0 & 96.0 & 94.0 & 56.0 & 68.0 & 66.0 & 66.0 & 74.0 & 0.0 \\
Yelp & 98.0 & 98.0 & 98.0 & 94.0 & 94.0 & 96.0 & 96.0 & 98.0 & 98.0 & 86.0 & 90.0 & 86.0 & 84.0 & 80.0 & 0.0 \\
IMDB & 96.0 & 98.0 & 100.0 & 96.0 & 96.0 & 92.0 & 82.0 & 96.0 & 96.0 & 76.0 & 80.0 & 80.0 & 84.0 & 80.0 & 0.0 \\
sentiment140 & 68.0 & 70.0 & 74.0 & 70.0 & 70.0 & 54.0 & 58.0 & 68.0 & 74.0 & 62.0 & 62.0 & 66.0 & 62.0 & 60.0 & 2.0 \\
\hline
\multicolumn{12}{l}{\textbf{READING Comp.}}\\
MultiRC & 68.0 & 64.0 & 60.0 & 52.0 & 38.0 & 44.0 & 44.0 & 48.0 & 44.0 & 54.0 & 52.0 & 50.0 & 48.0 & 40.0 & 6.0 \\
SQuADv1 & 68.0 & 62.0 & 62.0 & 66.0 & 68.0 & 64.0 & 64.0 & 62.0 & 68.0 & 68.0 & 70.0 & 66.0 & 66.0 & 54.0 & 4.0 \\
SQuADv2 & 62.0 & 26.0 & 8.0 & 56.0 & 12.0 & 30.0 & 20.0 & 22.0 & 16.0 & 24.0 & 24.0 & 26.0 & 22.0 & 16.0 & 0.0 \\
OBQA & 82.0 & 78.0 & 68.0 & 68.0 & 58.0 & 64.0 & 60.0 & 78.0 & 66.0 & 62.0 & 64.0 & 66.0 & 60.0 & 40.0 & 0.0 \\
BoolQ & 84.0 & 82.0 & 74.0 & 60.0 & 60.0 & 68.0 & 70.0 & 80.0 & 76.0 & 74.0 & 68.0 & 76.0 & 70.0 & 72.0 & 6.0 \\
Drop & 40.0 & 18.0 & 16.0 & 8.0 & 6.0 & 14.0 & 12.0 & 18.0 & 14.0 & 10.0 & 8.0 & 8.0 & 8.0 & 22.0 & 0.0 \\
\hline
\multicolumn{12}{l}{\textbf{CLOSE-BOOK QA}} \\
NQ & 18.0 & 12.0 & 10.0 & 16.0 & 10.0 & 16.0 & 14.0 & 16.0 & 10.0 & 12.0 & 12.0 & 12.0 & 4.0 & 12.0 & 0.0 \\
ARC-e & 50.0 & 74.0 & 80.0 & 56.0 & 70.0 & 54.0 & 56.0 & 66.0 & 82.0 & 58.0 & 58.0 & 60.0 & 58.0 & 48.0 & 0.0 \\
ARC-c & 46.0 & 46.0 & 48.0 & 42.0 & 46.0 & 34.0 & 34.0 & 50.0 & 46.0 & 46.0 & 42.0 & 42.0 & 42.0 & 24.0 & 0.0 \\
TriviaQa & 66.0 & 56.0 & 56.0 & 46.0 & 46.0 & 60.0 & 46.0 & 48.0 & 56.0 & 46.0 & 42.0 & 46.0 & 24.0 & 42.0 & 4.0 \\
\hline
\multicolumn{12}{l}{\textbf{COREFERENCE}}\\
DPR & 54.0 & 64.0 & 46.0 & 50.0 & 50.0 & 56.0 & 60.0 & 68.0 & 56.0 & 64.0 & 60.0 & 62.0 & 62.0 & 46.0 & 2.0 \\
WSC & 50.0 & 50.0 & 56.0 & 50.0 & 42.0 & 38.0 & 46.0 & 58.0 & 42.0 & 58.0 & 58.0 & 52.0 & 54.0 & 40.0 & 0.0 \\
\hline
\multicolumn{12}{l}{\textbf{READ. COOMP. W/ COMMONSENSE}}  \\
CosmosQa & 68.0 & 74.0 & 58.0 & 68.0 & 34.0 & 46.0 & 32.0 & 50.0 & 46.0 & 44.0 & 46.0 & 44.0 & 38.0 & 14.0 & 6.0 \\
record & 70.0 & 54.0 & 30.0 & 70.0 & 26.0 & 24.0 & 6.0 & 42.0 & 34.0 & 18.0 & 12.0 & 14.0 & 8.0 & 14.0 & 0.0 \\
\hline
\multicolumn{12}{l}{\textbf{PARAPHRASE}}    \\
Paws Wiki & 90.0 & 54.0 & 44.0 & 64.0 & 40.0 & 44.0 & 42.0 & 56.0 & 46.0 & 56.0 & 50.0 & 48.0 & 54.0 & 60.0 & 2.0 \\
QQP & 74.0 & 88.0 & 64.0 & 74.0 & 68.0 & 66.0 & 60.0 & 80.0 & 58.0 & 50.0 & 40.0 & 36.0 & 28.0 & 54.0 & 0.0 \\
MRPC & 60.0 & 62.0 & 60.0 & 58.0 & 58.0 & 60.0 & 62.0 & 60.0 & 58.0 & 42.0 & 44.0 & 40.0 & 42.0 & 60.0 & 2.0 \\
STSB & 38.0 & 34.0 & 8.0 & 36.0 & 16.0 & 12.0 & 12.0 & 30.0 & 20.0 & 20.0 & 20.0 & 20.0 & 14.0 & 12.0 & 0.0 \\
\hline
\multicolumn{12}{l}{\textbf{NLI}}\\
CB & 88.9 & 82.2 & 66.7 & 80.0 & 62.2 & 77.8 & 57.8 & 86.7 & 66.7 & 68.9 & 64.4 & 68.9 & 62.2 & 55.6 & 13.3 \\
WNLI & 70.0 & 52.0 & 42.0 & 68.0 & 46.0 & 44.0 & 50.0 & 60.0 & 54.0 & 56.0 & 56.0 & 42.0 & 44.0 & 52.0 & 0.0 \\
ANLI-r1 & 50.0 & 52.0 & 48.0 & 50.0 & 50.0 & 40.0 & 42.0 & 40.0 & 42.0 & 40.0 & 40.0 & 36.0 & 38.0 & 38.0 & 24.0 \\
ANLI-r3 & 46.0 & 50.0 & 38.0 & 42.0 & 38.0 & 38.0 & 40.0 & 44.0 & 50.0 & 28.0 & 32.0 & 34.0 & 38.0 & 40.0 & 24.0 \\
ANLI-r2 & 46.0 & 48.0 & 46.0 & 46.0 & 46.0 & 32.0 & 36.0 & 46.0 & 46.0 & 40.0 & 36.0 & 38.0 & 32.0 & 46.0 & 20.0 \\
ANLI-m & 88.0 & 76.0 & 84.0 & 84.0 & 88.0 & 62.0 & 66.0 & 80.0 & 88.0 & 48.0 & 54.0 & 50.0 & 56.0 & 76.0 & 0.0 \\
ANLI-mm & 92.0 & 84.0 & 90.0 & 90.0 & 94.0 & 64.0 & 82.0 & 88.0 & 90.0 & 48.0 & 48.0 & 50.0 & 60.0 & 84.0 & 2.0 \\
SNLI & 96.0 & 88.0 & 86.0 & 84.0 & 84.0 & 56.0 & 58.0 & 90.0 & 92.0 & 54.0 & 52.0 & 54.0 & 54.0 & 82.0 & 0.0 \\
QNLI & 94.0 & 66.0 & 62.0 & 94.0 & 26.0 & 46.0 & 48.0 & 74.0 & 38.0 & 56.0 & 56.0 & 54.0 & 60.0 & 70.0 & 0.0 \\
RTE & 52.0 & 68.0 & 64.0 & 62.0 & 72.0 & 54.0 & 58.0 & 70.0 & 76.0 & 64.0 & 58.0 & 56.0 & 64.0 & 80.0 & 22.0 \\
\bottomrule
\end{tabular}
}
\label{tab:full_results_13b}
\end{table*}

\begin{figure*}
    \centering
    \includegraphics[width=.8\linewidth]{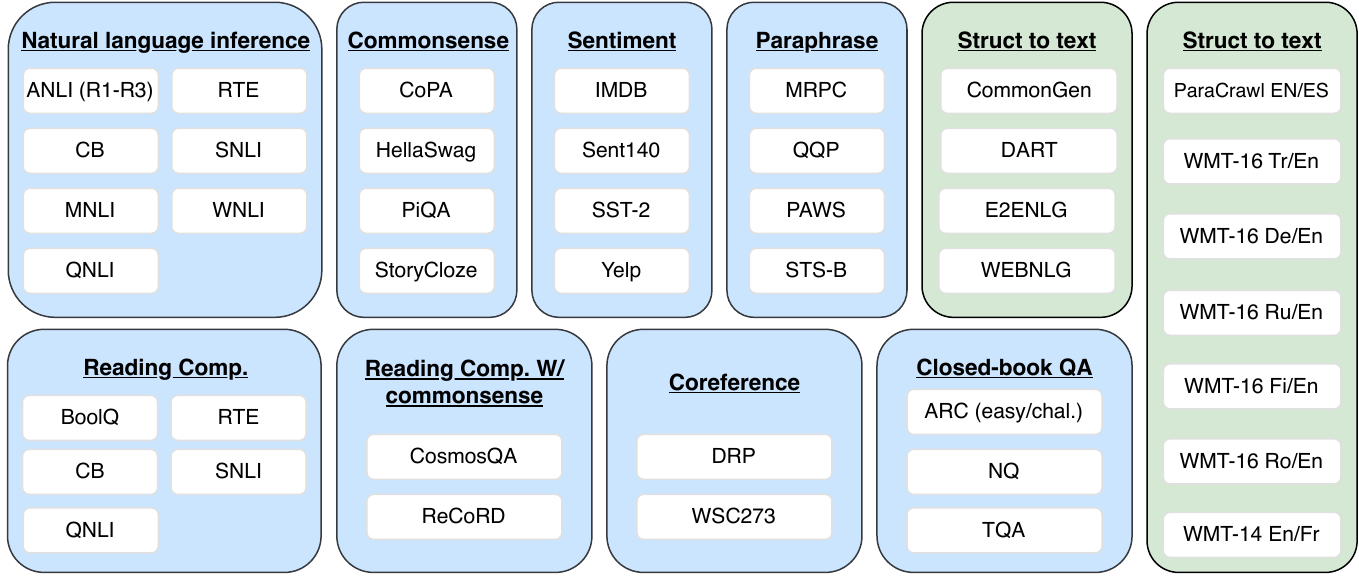}
    \caption{Datasets and task clusters used to train LoRAs and generate mixed-task evaluation set in this paper (NLU tasks in blue; NLG tasks in green).
    }
    \label{fig:tasks}
\end{figure*}
\section{Details of Training and Evaluation datasets}
\label{sec:eval_dataset}
We leverage a subset of flan-v2 datasets ~\cite{wei2021finetuned} as shown in Fig.\ref{fig:tasks} for LoRA expert training and mixed-task dataset generation. We summarize the details of the used datasets as follows:

\textbf{Struct-to-Text Conversion}: This task evaluates the capability to generate natural language descriptions from structured data inputs. We use the following datasets: (1) CommonGen; (2) DART; (3) E2ENLG; (4) WebNLG;

\textbf{Translation}: Translation involves converting text from one language to another, maintaining the original meaning and nuances. We use the following datasets: (1) En-Fr from WMT'14; En-De, En-Tr, En-Ru, En-Fi, En-Ro from WMT'16; (3) En-Es from Paracrawl.

\textbf{Commonsense Reasoning}: This involves assessing the ability to apply physical or scientific principles alongside common sense in reasoning tasks. We use the following datasets: (1) COPA, (2) HellaSwag, (3) PiQA, and (4) StoryCloze.

\textbf{Sentiment Analysis}: A fundamental task in natural language processing (NLP) that determines the sentiment polarity (positive or negative) of a given text. We use the following datasets: (1) IMDB, (2) Sentiment140, (3) SST-2, and (4) Yelp.

\textbf{Closed-Book Question Answering}: This task challenges models to answer questions about general knowledge without direct access to external information sources. We use the following datasets: (1) ARC, (2) NQ, and (3) TriviaQA.

\textbf{Paraphrase Detection}: This task requires models to ascertain whether two sentences convey the same meaning, indicating semantic equivalence. We use the following datasets: (1) MRPC, (2) QQP, and (3) Paws Wiki.

\textbf{Coreference Resolution}: Involves identifying instances within a text that refer to the same entity, demonstrating an understanding of textual context. We use the following datasets: (1) DPR and (2) WSC273.

\textbf{Reading comprehension}: Assesses the capability to derive answers to questions from a provided text containing relevant information. We use the following datasets: (1) BoolQ, (2) DROP, (3) MultiRC, (4) OBQA, (5) SQuADv1, (6) SQuADv2.

\textbf{Reading Comprehension with Commonsense}: Merges traditional reading comprehension skills with commonsense reasoning, requiring understanding beyond the explicit text. We use the following datasets: (1) CosmosQA; (2) ReCoRD.

\textbf{Natural Language Inference}: Focuses on deducing the relationship between two sentences, determining if the second sentence logically follows from, contradicts, or is unrelated to the first sentence. We use the following datasets: (1) ANLI, (2) CB; (3) MNLI; (4) QNLI; (5) SNLI; (6) WNLI; (7) RTE.

\end{document}